**Nonlinear second-order dynamics describe labial constriction trajectories across languages and contexts**

Michael C. Stern* & Jason A. Shaw

Department of Linguistics, Yale University

*Corresponding author address: 370 Temple St, New Haven, CT 06511

*Corresponding author email: michael.stern@yale.edu

**Acknowledgements:** We would like to thank Ben Kramer and Yichen Wang for their work on data collection and processing. Aspects of the data described in this paper were first reported at the third Hanyang International Symposium on Phonetics and Cognitive Sciences of Language (HISPhonCog) and the 20th International Congress of Phonetic Sciences (ICPhS). A preliminary version of the model was presented at the 13th International Seminar on Speech Production (ISSP). We would like to thank audiences at each of these meetings for their feedback.

**Declaration of Interests:** The authors declare that they have no known competing financial interests or personal relationships that could have appeared to influence the work reported in this paper.

**Abstract**

We investigate the dynamics of labial constriction trajectories during the production of /b/ and /m/ in English and Mandarin in two prosodic contexts. We find that, across languages and contexts, the ratio of instantaneous displacement to instantaneous velocity generally follows an exponential decay curve from movement onset to movement offset. We formalize this empirical discovery in a differential equation and, in combination with an assumption of point attractor dynamics, derive a nonlinear second-order dynamical system describing labial constriction trajectories. The equation has only two parameters, $T$ and $r$. $T$ corresponds to the target state and $r$ corresponds to movement rapidity. Thus, each of the parameters corresponds to a phonetically relevant dimension of control. Nonlinear regression demonstrates that the model provides excellent fits to individual movement trajectories. Moreover, trajectories simulated from the model qualitatively match empirical trajectories, and capture key kinematic variables like duration and peak velocity. The model constitutes a proposal for the dynamics of individual articulatory movements, and thus offers a novel foundation from which to understand additional influences on articulatory kinematics like prosody, inter-movement coordination, and stochastic noise.

**Highlights:**

- Ratio of displacement to velocity in lip constrictions follows exponential decay.
- A nonlinear second-order dynamical model derives this observation.
- Model parameters are $T$ (target state) and $r$ (movement rapidity).
- Model provides excellent fits to acceleration as a function of velocity and state.
- Model-generated trajectories capture key measures of articulatory kinematics.

## 1. Introduction

### *1.1 Background*

Speech articulation involves coordinated movements of articulatory organs like the tongue, lips, and larynx in order to generate sound. These movements can be characterized as dynamical systems: sets of variables which change lawfully over time. The formal expression of a dynamical system is given in Eq. 1:

$$\dot{\boldsymbol{x}} = \boldsymbol{f}(\boldsymbol{x}) \tag{1}$$

$\boldsymbol{x}$ is the set of variables which characterizes the state of the system at any given time. In speech articulation, this set of variables could in theory be extremely large, e.g., including the three-dimensional spatial position of each articulator, the tension of each muscle relevant for moving the articulators, or even the membrane potential of each neuron relevant for articulatory motor control. In practice, however, speech articulation can be described using a relatively small number of "task" variables which are subserved by flexible synergies of other variables, as in Task Dynamics (Saltzman & Munhall, 1989). In this way, the high-dimensional space of potentially relevant variables is reduced. For instance, *lip aperture* (LA)—the distance between the upper and lower lip—is a task variable relevant for producing labial sounds like /b/ and /m/, which is controlled by a synergy between the positions of the upper lip, lower lip, and jaw.

Eq. 1 describes the time derivatives of $\boldsymbol{x}$, i.e., $\dot{\boldsymbol{x}}$ or the rates of change of $\boldsymbol{x}$, as some set of functions $\boldsymbol{f}(\boldsymbol{x})$. An important benefit of such a dynamical characterization of speech articulation is that it allows an explicit link between time-variant articulatory trajectories and stable (relatively time-invariant) control parameters, i.e., the coefficients of the terms in $\boldsymbol{f}(\boldsymbol{x})$. These control parameters can be interpreted as cognitive representations. For instance, Articulatory Phonology (e.g., Browman & Goldstein, 1989) links control parameters of the functions in Task Dynamics (in particular, "stiffness" $k$ and target position $T$) to dimensions of phonological contrast in language. In this way, dynamical models bridge the traditional divide between discrete and continuous representations in linguistics (Iskarous, 2017; Iskarous & Pouplier, 2022) and cognitive science more broadly (van Gelder, 1998). In order for the control parameters of $\boldsymbol{f}(\boldsymbol{x})$ to be theoretically interpretable, they should be relatively few in number (e.g., Brunton et al., 2016) while still capturing key properties of articulation. The importance of minimizing control parameters in the dynamics is particularly relevant for making a principled theoretical connection between articulation and phonology, a low-dimensional combinatoric system.

A useful kind of dynamical system for modeling speech articulation (as well as other kinds of goal-directed behavior) is a *point attractor*, as in Eq. 2.[1]

$$\lambda \dot{x} = -x + T \tag{2}$$

The system described by Eq. 2 has a point attractor because $\dot{x}$ is negatively related to $x$. Regardless of the particular value of $x$ at any given time, $\dot{x}$ will be in the direction of $T$, i.e., $x$ will be moving towards $T$. Moreover, when $x = T$, $\dot{x} = 0$. In other words, $x$ stops moving when $x$ reaches $T$. For this reason, $T$ is the point attractor of the system, called $T$ in this case for "target" (cf. Mücke et al., 2024). Moreover, the magnitude of $\dot{x}$ at any given time is proportional to the difference between $x$ and $T$ at that time, as well as the control parameter $\lambda$. In other words, the farther away $x$ is from $T$, the more quickly $x$ moves towards $T$, with the actual magnitude modulated by $\lambda$. This captures an important fact about speech articulation, namely that the peak velocity of a movement is robustly correlated with total movement amplitude (Ostry & Munhall, 1985). The farther an articulator travels to reach its target, the faster it moves, although the details of this relationship are modulated by phonological category, prosodic prominence, and speech rate (Beckman & Edwards, 1992; Kelso et al., 1985; Ostry et al., 1983).

Speech articulation can be characterized as a sequence of target-directed movements governed by point attractor dynamics. For instance, articulating /b/ requires bringing the lips together to form a closure. In this case, $x$ is lip aperture (LA), $\lambda$ controls the rate of movement, and $T$ is either zero, indicating that the lips are together, or a small negative number, indicating a target just beyond what is physically possible (e.g., Löfqvist & Gracco, 1997; Parrell, 2011). In this way, Eq. 2 captures the basic target-directedness of articulatory movement, with differences between movements arising from variation in the positions of the targets and in the speeds of the movements. However, the shapes of the trajectories generated by Eq. 2 are quite different from observed trajectories. In particular, for any fixed value of the control parameter $\lambda$, simulated trajectories achieve peak velocity instantaneously; velocity then decreases monotonically as $x$ approaches $T$. In real trajectories, peak velocity occurs later, approximately halfway through the movement (Ostry et al., 1987). The observed delay in the achievement of peak velocity has inspired the use of a *damped mass-spring* model of speech articulation, as in Eq. 3:

$$b\dot{x} = k(-x + T) - m\ddot{x} \tag{3}$$

[1] Henceforth we focus on the dynamics of individual variables, hence the unbolded symbols.

In Eq. 3, peak velocity is delayed (relative to Eq. 2) because velocity $\dot{x}$ is negatively related to acceleration $\ddot{x}$. This empirical improvement is achieved via greater model complexity: Eq. 3 is a second-order system, referencing the second time derivative $\ddot{x}$ or acceleration, unlike the first-order system in Eq. 2, which only references the first time derivative $\dot{x}$ or velocity. Moreover, Eq. 3 has four control parameters $m$, $b$, $k$, and $T$, more than the two parameters $\lambda$ and $T$ in Eq. 2.[2]

Even in the second-order model in Eq. 3, however, peak velocity occurs unrealistically early (Perrier et al., 1988). Thus, more complexity has been added to Eq. 3. One approach has been to add a time-varying activation parameter $a(t)$ (Byrd & Saltzman, 1998; Kröger et al., 1995). Allowing $a(t)$ to ramp up over time successfully delays the achievement of peak velocity, more closely matching empirical velocity trajectories. Another approach has been to add a cubic term $d(-x+T)^3$ (Sorensen & Gafos, 2016). This delays the achievement of peak velocity by reducing acceleration at large distances from the target. This approach also captures nonlinearity in the relationship between peak velocity and total movement amplitude (Ostry & Munhall, 1985). Both of these approaches achieve greater empirical adequacy, but at a cost. The first approach introduces a dependence on time $t$, making the system *non-autonomous*, which has been argued to be undesirable for dynamical models of behavior (Fowler, 1980; Sorensen & Gafos, 2016). The second approach preserves autonomy, but only by introducing an additional parameter $d$, increasing model complexity and reducing the theoretical interpretability of each parameter.[3]

### *1.2 Empirical bases of existing dynamical models*

In this subsection we summarize the data that has informed the existing dynamical models of articulation described in Section 1.1, in order to better situate the data reported in the present study. One of the first applications of the damped mass-spring model to speech articulation was from Ostry & Munhall (1985), who adapted the idea from research on limb movements (Cooke, 1980). Ostry & Munhall (1985) used ultrasound imaging to measure tongue dorsum kinematics during repetition of CV syllables from three speakers of Canadian English. They found a robust correlation between the maximum spatial displacement of a movement and its peak velocity in all three speakers, which is well-captured by the damped mass-spring model. They also proposed that the model's stiffness parameter $k$ is involved in controlling speech rate. Kröger et al. (1995) enriched the damped mass-spring model by introducing continuous activation ramping and de-ramping at the beginning and end of movements, respectively. They used electromagnetic articulography (EMA) to track lip, tongue tip, and tongue body movements of three native speakers of German during production of CV syllables which were parts of real words in carrier phrases. Addition of the activation-ramping mechanism was largely motivated by symmetrical velocity shapes, which cannot be generated by the original damped mass-spring model. For similar reasons, Byrd & Saltzman (1998) also used activation ramping in their implementation of the damped mass-spring model. Their data came from EMA recordings of lip movements in three speakers of English producing sentences with target words in varying prosodic contexts. They found that prosodic variation could be captured through a combination of variation in stiffness $k$ and variation in parameters controlling the shape of the activation curve. Sorensen & Gafos (2016) replaced activation ramping with a cubic term in order to eliminate explicit dependence on time and therefore preserve system autonomy. They analyzed two sets of data: one dataset consisting of x-ray microbeam (XRMB) recordings of tongue dorsum movements from 43 English speakers producing isolated nonwords, and another consisting of EMA recordings of lip movements from three English speakers producing repeated CV syllables. They found that the damped mass-spring model with an additional cubic term could generate the observed symmetrical velocity trajectories and cubic-shaped Hooke diagrams (acceleration by displacement) without the need for time-dependent activation ramping. Finally, Kuberski & Gafos (2023) evaluated fits of the original damped mass-spring model to EMA recordings of tongue tip and tongue body movements during repetition of CV syllables at different metronome rates in five speakers of English and five speakers of German. They found that, in general, fits were better at faster rates and with more aggressive thresholding for the measurement of movement onsets and offsets, reflecting the fact that the relationship between acceleration and position is less linear at greater spatial displacement.

---

[2] Some implementations of the damped mass-spring model constrain certain parameters so that there are functionally fewer than four free parameters. Most commonly, $m = 1$ and $b = 2\sqrt{k}$ in order ensure "critical" damping or point attractor behavior (Ostry & Munhall, 1985; Saltzman & Munhall, 1989).

[3] Another theoretical approach that improves model fits to articulatory data relative to the original damped mass-spring model is General Tau theory (Lee, 1998), as developed in Elie et al. (2023, 2024). General Tau theory describes point attractor dynamics that are both non-autonomous and nonlinear, and require reference to the temporal duration of the movement. This last feature brings it beyond the incremental theoretical development pursued in this paper.

### *1.3 This study*

The current study has two related goals. The first is to extend the body of data bearing on speech articulatory dynamics by examining a relatively large number of subjects (24) producing meaningful sentences in two unrelated languages in two different prosodic contexts. We varied prosodic context in order to elicit a greater degree of articulatory variability, e.g., in speed and duration (e.g., Byrd & Saltzman, 1998). Additionally, we included languages with different prosodic profiles to magnify the variability introduced by prosody. Our prosodic manipulation varies the position of focus in the target sentences, which is known to have effects on word duration in both English (~100 ms, Cooper et al., 1985) and Mandarin (~30-50 ms, Xu, 1999). Our goal is not to compare the results between different conditions (language or prosodic context) but rather to increase the variability in the data. This facilitates our second goal, which is to derive a maximally generalizable dynamical model. Our approach is to utilize the natural variability in the data to constrain model discovery. This differs from another valid scientific approach of starting with more idealized data (less variability) to discover a basic model which can be elaborated as more variability is considered.

In this study, we advance a method for uncovering the basic dynamics of articulatory movements, with a focus on balancing model simplicity and empirical adequacy. Rather than starting from the specific second-order system in Eq. 3, we start from a minimal assumption of point attractor dynamics, formalized in Eq. 2. This allows us to solve for the parameter $\lambda$ from measurement of data, in particular, electromagnetic articulography (EMA) recordings of bilabial constriction movements during the production of /b/ and /m/ in CV sequences in English and Mandarin speakers in two different prosodic contexts: words produced with and without contrastive focus. In this way, we address the question: what is the empirical relationship between velocity $\dot{x}$ and state $x$ over time? The answer to this question guides dynamical model development.

We focus on bilabial movements because the hypothesized task dimension of lip aperture (LA) is transparently related to the measurable spatial positions of the upper and lower lips (via the Euclidean distance). In contrast, task dimensions governing tongue movements, i.e., constriction location and constriction degree, are not as transparently related to measurable spatial positions of fleshpoints on the tongue. Moreover, we focus on the constriction phase of the movement (cf. Kuberski & Gafos, 2023), rather than the release phase, because there is evidence that the dynamics of the release phase are coupled to the following vowel constriction movement (Kramer et al., 2023). This coupling would introduce additional complexity into the kinematics of the movement. Thus, labial constriction movements present an ideal test case for uncovering the dynamics of individual articulatory movements. An important next step will be to extend the discovered dynamics to a wider range of movement types, e.g., consonant release movements and vowel constriction movements, in tandem with coordination dynamics.

## 2. Experimental methods

### *2.1 Data availability*

The data, analysis code, and simulation code for generating the results reported below are available on OSF at https://osf.io/7f2mh/.

### *2.2 Participants*

Data was collected from 24 subjects: 12 native speakers of American English (8 female, 4 male, ages 19–28, mean = 20.75) and 12 native speakers of Mandarin Chinese (7 female, 4 male, 1 nonbinary, ages 19–33, mean = 24.00). All participants self-reported no history of speech, language, or hearing impairment.

### *2.3 Materials*

Stimuli consisted of eight word-initial CV sequences in each language, where the initial consonant was bilabial—either [b] or [m]—and the vowel was either low back [ɑ] or high front [i]. Target sequences containing the vowel [i] were immediately preceded by the vowel [ɑ], and sequences containing the vowel [ɑ] were immediately preceded by the vowel [i]. All Mandarin target syllables bore a falling tone (T4) and were preceded immediately by a low tone (T3). Each target syllable was produced in two carrier sentences, occurring once in an informationally prominent position and once in a less prominent position. To encourage natural speech, each carrier sentence was preceded by a question, which served to provide context for the target sentences. Examples of context questions and target carrier sentences are given in Table 2. A full list of stimuli is included in Appendix B.

| Item | Language | Prominence | Stimulus | |
|---|---|---|---|---|
| [imɑ] | English | Not prominent | Prompt | Is she a knee model client? |
| | | | Target | She's a knee model **representative**, not a knee model client. |
| | | Prominent | Prompt | Is she a knee surgeon? |
| | | | Target | She's a knee **model**, not a knee surgeon. |
| | Mandarin | Not prominent | Prompt | Wo3 ying1gai1 ma4 ta1 hai2shi4 ma4 ni3?<br>'Should I scold him or scold you?' |
| | | | Target | Ni3 ma4 **ta1** jiu4 xing2 le0, bie2 ma4 wo3.<br>'Just scold him; don't scold me.' |
| | | Prominent | Prompt | Wo3 ying1gai1 ma4 ta1 hai2shi4 da3 ta1?<br>'Should I scold him or hit him?' |
| | | | Target | Ni3 **ma4** ta1 jiu4 xing2 le0, bie2 da3 ta1.<br>'Just scold him; don't hit him.' |

Table 2: Example stimuli. Target CV sequences are underlined; prominent words are bolded.

### 2.4 Procedure

Presentation of materials was controlled using E-Prime. On each trial, an audio recording of a question was played. The question was also displayed in text on the screen for 5000 ms. Participants were instructed to listen to the question and to read aloud the answer that followed. In total, each participant produced 128 tokens (8 items × 2 carrier sentences × 8 repetitions) across four blocks of 32 items each. Within each block, stimuli were presented in a randomized order.

Articulatory kinematic data was collected with the NDI Wave Speech Research System sampling in five dimensions (three positional dimensions and two angular dimensions) at a rate of 100 Hz. The sensors of interest for this study were attached at the vermilion border of the upper lip (UL) and lower lip (LL). Three sensors were also attached to the tongue: tongue tip (TT), tongue blade (TB), and tongue dorsum (TD), placed ~1 cm, ~3 cm, and ~5 cm from the tip of the tongue, respectively. In order to track movements of the jaw, one lower incisor (LI) sensor was attached to the hard tissue of the gum directly below the left incisor. Reference sensors were attached on the left and right mastoids and on the nasion. Measurements of the occlusal plane and a midsagittal palate trace were also collected. The occlusal measurement was made by having the participant hold between their teeth a rigid planar object with three EMA sensors attached. Acoustic data was collected using a Sennheiser shotgun microphone at a sampling rate of 22,050 Hz.

### 2.5 Data processing

Articulatory data was rotated to the occlusal plane and corrected for head movement computationally. We followed a standard method, making use of our bite plane occlusal trial and reference sensor positions to calculate a rigid body rotation applied to the remaining sensors across trials. Trajectories were smoothed using the robust smoothing algorithm of Garcia (2010), *smoothn* function in MATLAB. First and second time derivatives (velocity and acceleration) were calculated from the smoothed trajectory using central differencing. All trajectories were upsampled to 1000 Hz using the *resample* function in MATLAB in order to improve the temporal precision of movement landmarks. Consonant constriction movements were parsed from the lip aperture (LA) signal, calculated as the Euclidean distance between the UL and LL sensors. The onset and offset of each movement were marked as the timepoints at which velocity exceeded or fell below, respectively, a 20% threshold of peak velocity.[4] Onsets and offsets were first semi-manually labeled using

[4] Past work has shown that the fit of the damped mass-spring model to kinematic data is sensitive to the threshold of peak velocity used for the movement parse (Kuberski & Gafos, 2023). For this reason, we also parsed our trajectories at 10% of peak velocity. Ultimately, the general pattern of results described in Sections 3 and 4—the

the *findgest* procedure in the MATLAB-based software *mview* (Tiede, 2005). After upsampling, the original landmarks were automatically adjusted to correct landmarking errors. This procedure involved finding the peak velocity within the timeframe delimited by the original landmarks, and then searching outward from the timepoint of peak velocity to find the timepoints (before and after peak velocity) at which the threshold was reached. Figure 1 displays an example LA trajectory with onset and offset landmark labels.

---

shapes of the average trajectories (empirical and simulated), the distributions of kinematic measures (empirical and simulated), and the correlations between empirical and simulated measures—was unaffected by the choice of threshold. Including more of the trajectory (10% threshold) led to more data exclusions because of non-monotonicity in the trajectory (458 exclusions with the 10% threshold compared to 87 exclusions with the 20% threshold; see the final paragraph of this section). We also found that model fit, on average, decreased slightly with the lower threshold (mean $R^2$ = 0.83 with the 10% threshold compared to mean $R^2$ = 0.87 with the 20% threshold; see Section 4.2), cf. Kuberski & Gafos (2023), who found that lower thresholds also tended to reduce the fit of the linear damped mass-spring model. Thus, the trend that lower thresholds reduce dynamical model fit appears to generalize beyond the linear damped mass-spring model. In the remainder of the paper, we focus on the trajectories parsed using the 20% threshold.

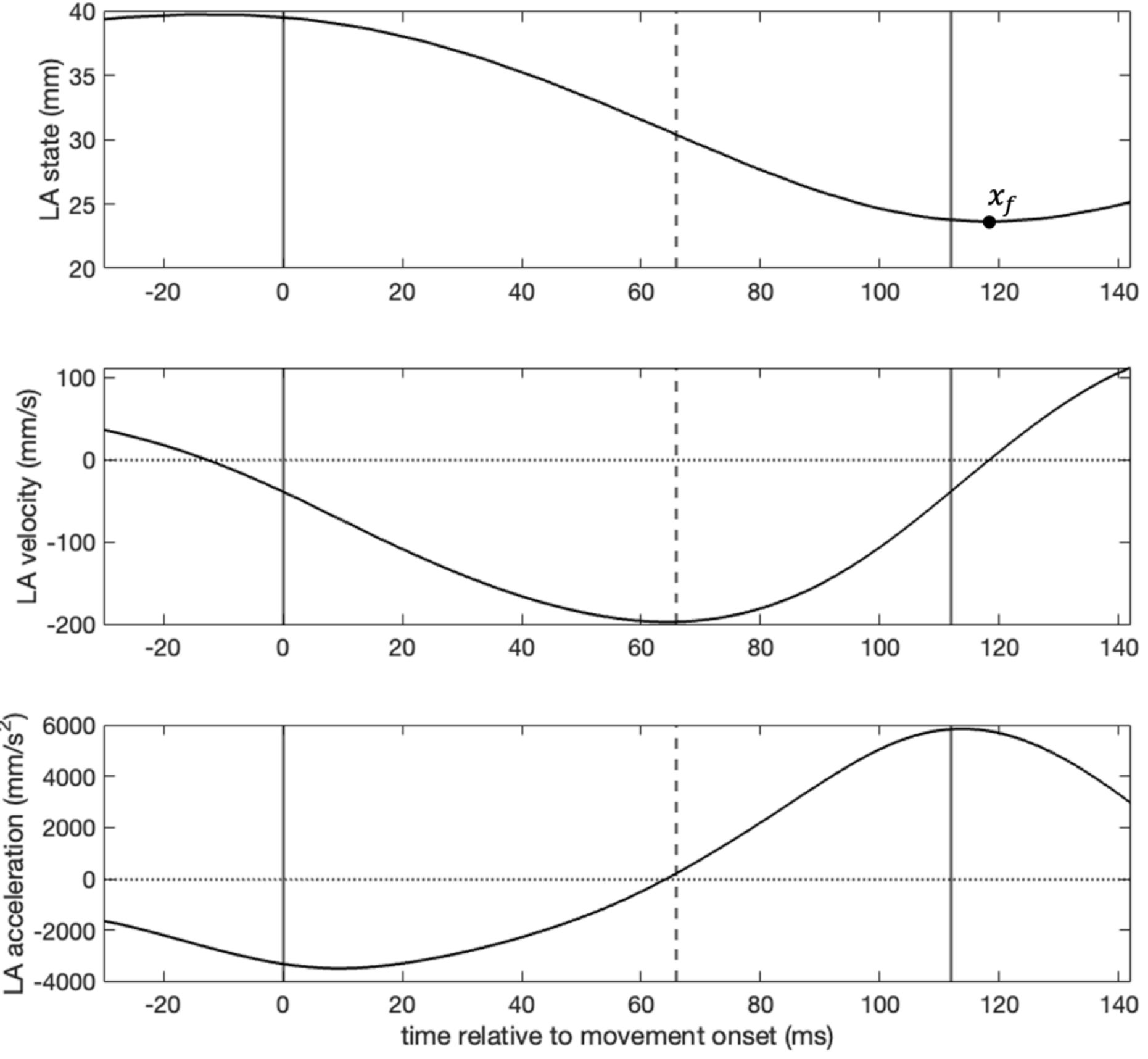

Figure 1: Example LA state (top), velocity (middle), and acceleration (bottom). The dashed vertical line corresponds to the timepoint of peak velocity. The solid vertical lines correspond to the timepoints at which velocity exceeded or fell below 20% of peak velocity. These landmarks define the onset and offset of movement. The dotted horizontal lines in the velocity and acceleration plots correspond to zero. LA state at the timepoint of the velocity zero-crossing, $x_f$, is labeled with a dot.

We calculated displacement at each sample as the difference between the current state of LA $x$ and the state of LA at the timepoint of the velocity zero-crossing following movement offset $x_f$, i.e., $x_f - x$ (cf. $-x + T$ as in Eq. 2). $\lambda$ was calculated at each sample as the ratio of instantaneous displacement to instantaneous velocity: $(-x + x_f)/\dot{x}$ (see Eq. 2). By demarcating movements based on a 20% threshold of peak velocity, instead of, e.g., velocity zero-crossing, we exclude portions of the kinematics in which displacement or velocity are infinitesimal. This facilitates the calculation of $\lambda$ from data, as it prevents $\lambda$ from approaching 0 (infinitesimal displacement) or infinity (infinitesimal velocity). Movement duration was calculated by subtracting the timestamp of the onset of movement from the timestamp of the offset of movement. We also calculated a measure of "kinematic stiffness" for each movement by dividing peak velocity by maximum displacement (e.g., Roon et al., 2021). This measure is a proxy for movement speed which accounts for the strong relation between peak velocity and maximum displacement. In this dataset, maximum displacement was always onset state minus final state. In order to plot trajectories together, all trajectories were normalized to 100 time units using shape-preserving cubic Hermite interpolation.

Out of the 3072 tokens elicited, a total of 675 tokens (22.0%) were eliminated from analysis for the following reasons: failure of the parsing tool to extract the movement (379 tokens); a non-monotonic trajectory, i.e., instantaneous velocity changed sign for at least one sample (87 tokens); failure of the participant to produce contrastive focus on the informationally prominent syllable, as judged by a research assistant (155 tokens); disfluency (5 tokens); or data storage failure (49 tokens). 2398 tokens, all of which involved monotonic decreases in distance to target, entered the analysis.

## 3. Experiment results

### *3.1 Kinematics*

Figure 2 displays average trajectories of displacement (left), velocity (center), and acceleration (right) over time, from movement onset to movement offset, as defined in Section 2.5 (see Figure 1).

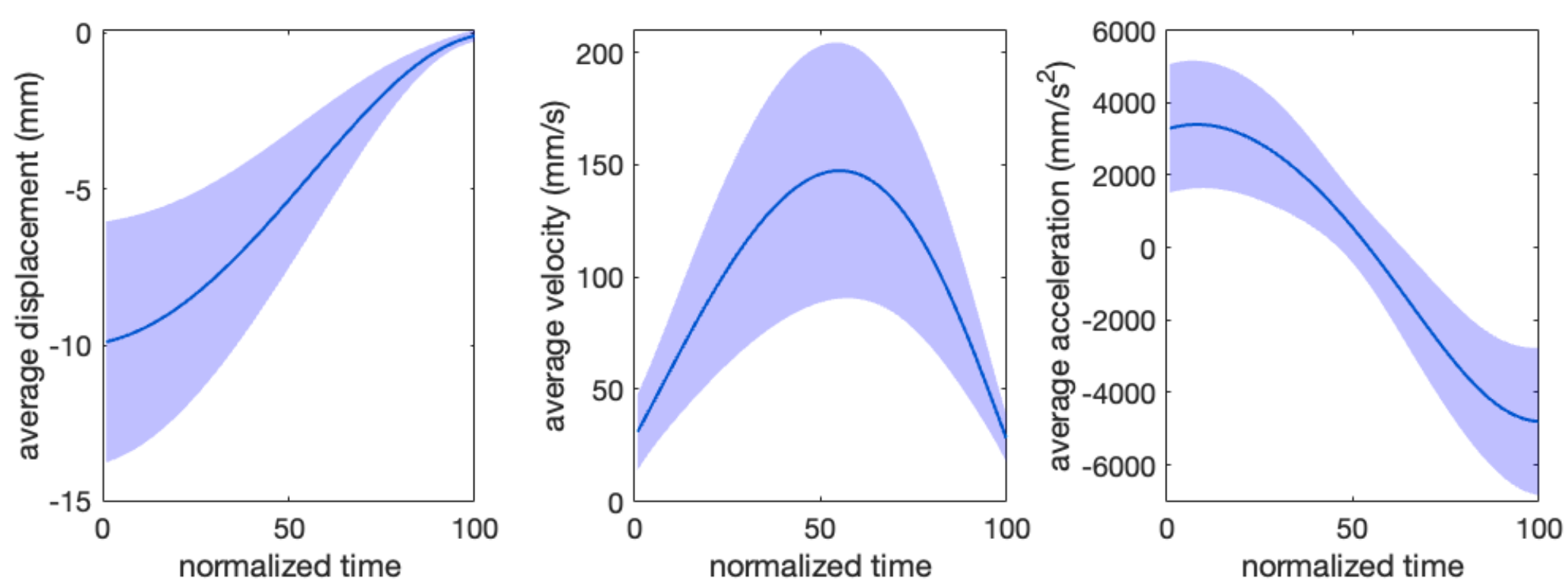

Figure 2: Average trajectories across all tokens ($n$ = 2398). Shading indicates one standard deviation. Note that these trajectories are calculated over *displacement* (distance between current LA state and LA state at minimum velocity following movement offset), rather than LA state itself, as in Figure 1.

It can be seen that displacement follows a sigmoidal shape, corresponding to a parabolic shape in the velocity curve and a cubic shape in the acceleration curve. Figure 3 displays the distributions of the kinematic variables duration, maximum displacement, kinematic stiffness, peak velocity, absolute time to peak velocity, and relative time to peak velocity (as a ratio of movement duration), across all 2398 tokens from all 24 speakers.

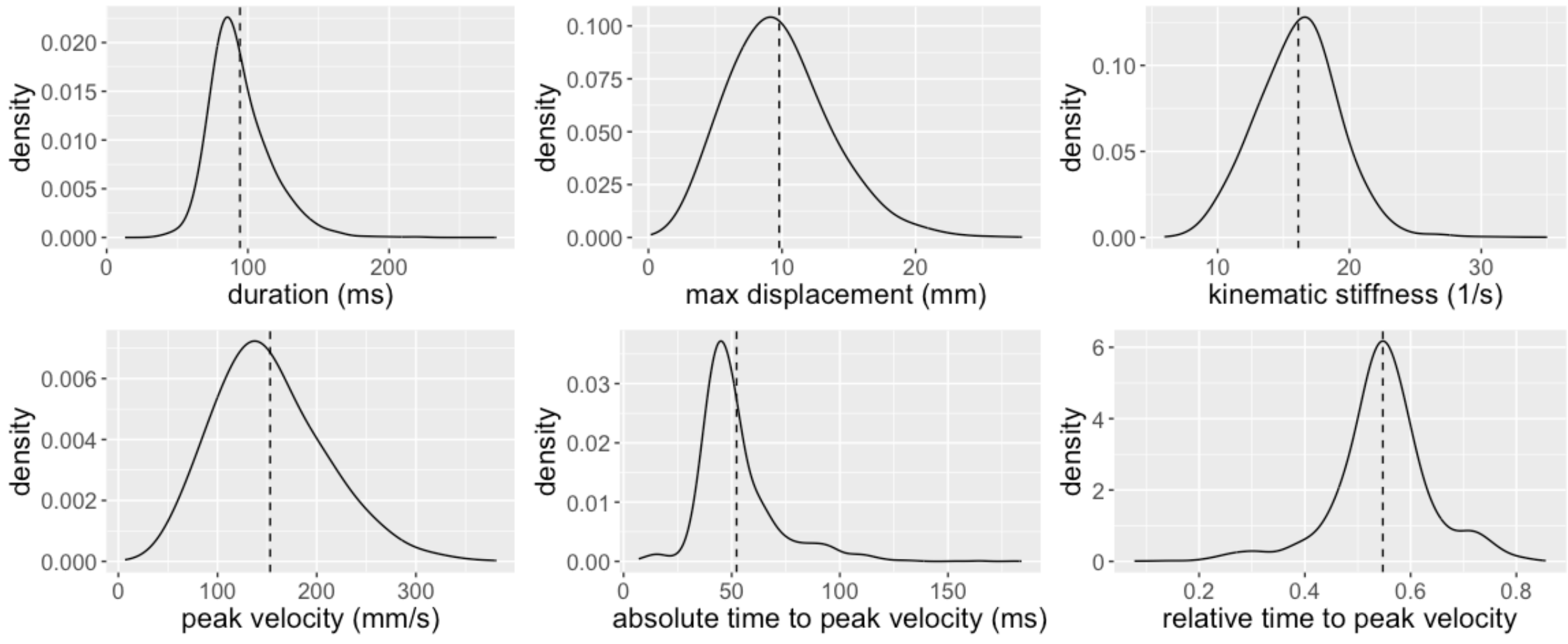

Figure 3: Distributions of kinematic variables across all tokens (*n* = 2398). Dashed vertical lines indicate the mean.

Mean movement duration was 94.39 ms, and the distribution of measurements had a long rightward tail (Figure 3: top left). The distribution of the other durational measure, absolute time to peak velocity, also had a long rightward tail, with a mean of 52.25 ms (Figure 3: bottom center). The other distributions are more normal. The average trajectories in Figure 2 and the distributions of kinematic variables in Figure 3 provide two distinct but related perspectives on the data. Consider the timing of the peak in the average velocity curve in Figure 2 in tandem with the distribution of relative time to peak velocity values in Figure 3. The peak in the average velocity curve is at 0.55 of total movement duration. Correspondingly, mean relative time to peak velocity across all movements was 0.55. Thus, on average, the velocity curves in this dataset are almost symmetric, but peak velocity tends to occur slightly later than halfway through the movement. This represents an apparent contrast with previous studies that have found peak velocity tending to occur slightly earlier than halfway through the movement (e.g., Byrd & Saltzman, 1998; Ostry et al., 1987). However, these studies also discovered variation in the shapes of velocity curves as a function of prosodic context, speech rate, and individual. In our data, *average* relative time to peak velocity is 0.55, but there is substantial variation around this mean, including a large number of tokens (N = 505) where peak velocity occurs earlier than halfway through the movement. As discussed in Section 1.3, examining by-condition variation is beyond the scope of this study, where we aim to derive a dynamical model from data across a range of natural variation.

*3.2 λ trajectories*

Next we examine the trajectories of $\lambda$, i.e., the ratio of instantaneous displacement to instantaneous velocity. As seen in Figure 4 (left), $\lambda$ generally followed an exponential decay curve from movement onset to offset. Consistent with this observation, the natural log of $\lambda$, henceforth $\ln(\lambda)$, follows an approximately linear trend (Figure 4: right).

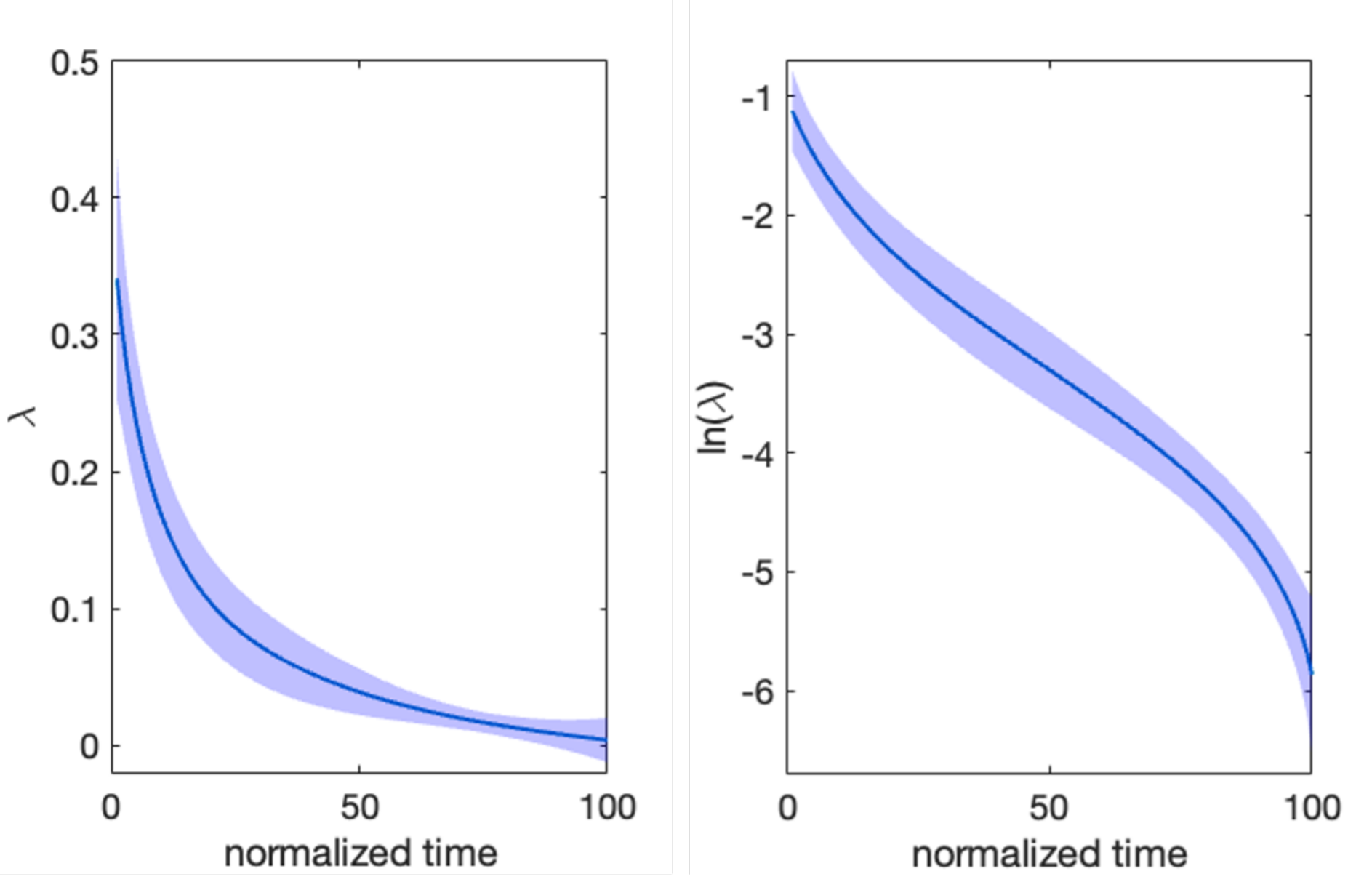

Figure 4: Average trajectories of $\lambda$ (left) and $\ln(\lambda)$ (right) across all tokens (*n* = 2398). Shading indicates one standard deviation.

To evaluate the robustness of this generalization, a linear regression model was fit to each trajectory of $\ln(\lambda)$ over time. The fits were excellent: overall mean $R^2 = .97$. Moreover, as seen in Figure 5, the slope of each linear fit correlates strongly with linguistically relevant measures like movement duration (Spearman's $\rho = .90$, $p < .001$) and kinematic stiffness ($\rho = -.91$, $p < .001$). Steeper (more negative) slopes correspond with decreased duration and increased kinematic stiffness. It is noteworthy that the relationship with duration appears less linear than the relationship with kinematic stiffness. This is likely because kinematic stiffness factors out some of the natural covariation between kinematic variables, and is thus conceptually closer to a dynamic control parameter than raw duration (e.g., Kelso et al., 1985).

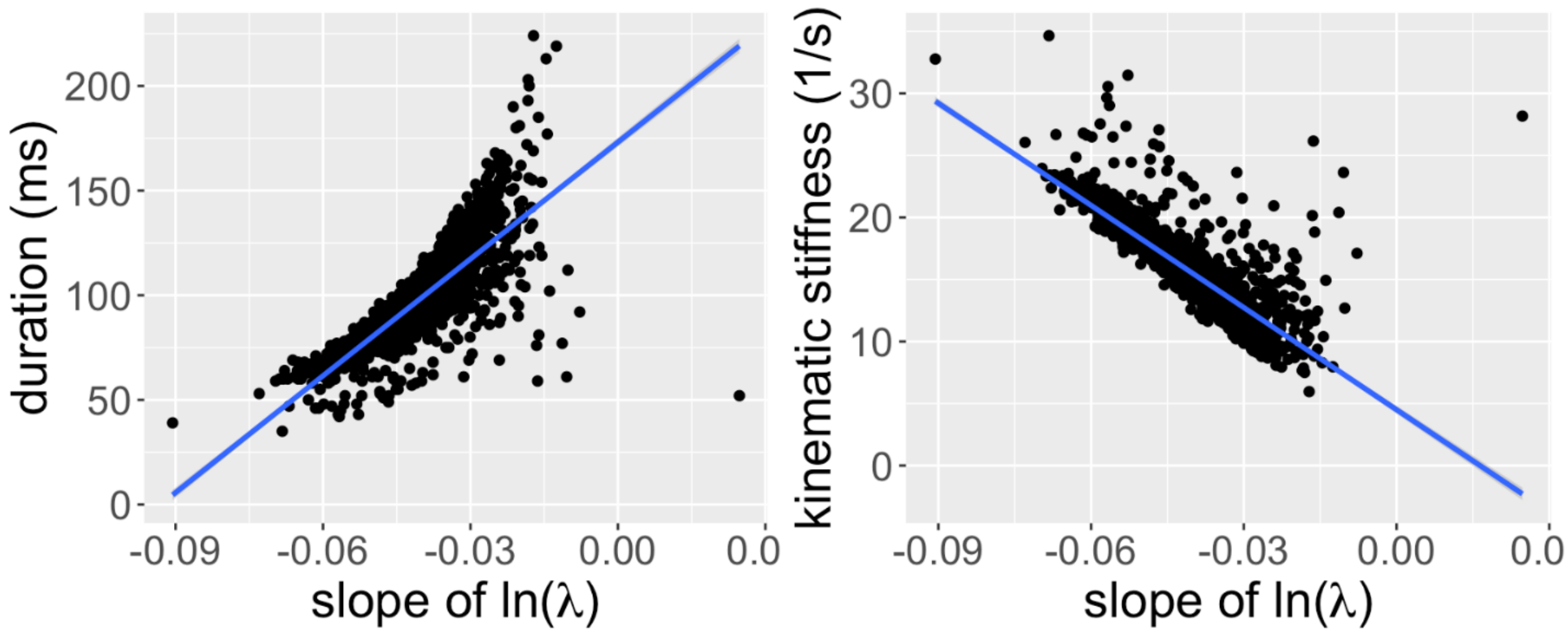

Figure 5: Correlations between the slope of a regression line fit to $\ln(\lambda)$ and two kinematic variables: movement duration (left) and kinematic stiffness (right).

## 4. Gestural model

### *4.1 Formalization*

The empirical observation of exponential decay in $\lambda$ over time can be expressed as a differential equation as in Eq. 4, with $r$ controlling the rate of decay. We call this parameter $r$ because the rate of decay in $\lambda$ corresponds to the "rapidity" of movement, as seen in Figure 5. Higher values of $r$ correspond to shorter duration (Figure 5: left) and higher kinematic stiffness (Figure 5: right).

$$\dot{\lambda} = -r\lambda \tag{4}$$

Together, the two first-order equations in Eq. 2 and Eq. 4 express a dynamical system of two variables, $x$ and $\lambda$. Since $\lambda$ is defined in Eq. 2 as $(-x + T)/\dot{x}$, we can substitute this definition into Eq. 4 to derive a single second-order equation, eliminating $\lambda$. This equation, solved for acceleration $\ddot{x}$, is shown in Eq. 5. See Appendix A for the full derivation.

$$\ddot{x} = r\dot{x} - \dot{x}^2/(-x + T) \tag{5}$$

Eq. 5 has only two parameters, $T$ and $r$, which can both be inferred from data and have clear interpretations. $T$ corresponds to the spatial target, and $r$ corresponds to movement rapidity, similar to stiffness $k$ in the damped mass-spring model. Moreover, the system is autonomous as it does not reference time (Fowler, 1980; Sorensen & Gafos, 2016). Unlike the linear damped mass-spring model, the system in Eq. 5 is nonlinear, as seen in the second term in the right side of the equation. This issue is discussed further in Section 5.

*4.2 Model fitting*

We used nonlinear least squares regression to fit the parameters $T$ and $r$ in Eq. 5 to each acceleration $\ddot{x}$ curve as a function of state $x$ and velocity $\dot{x}$ using the *fit* function in MATLAB. The distribution of $R^2$ values (indicating the goodness of fit) across the 2398 tokens is displayed in Figure 6.

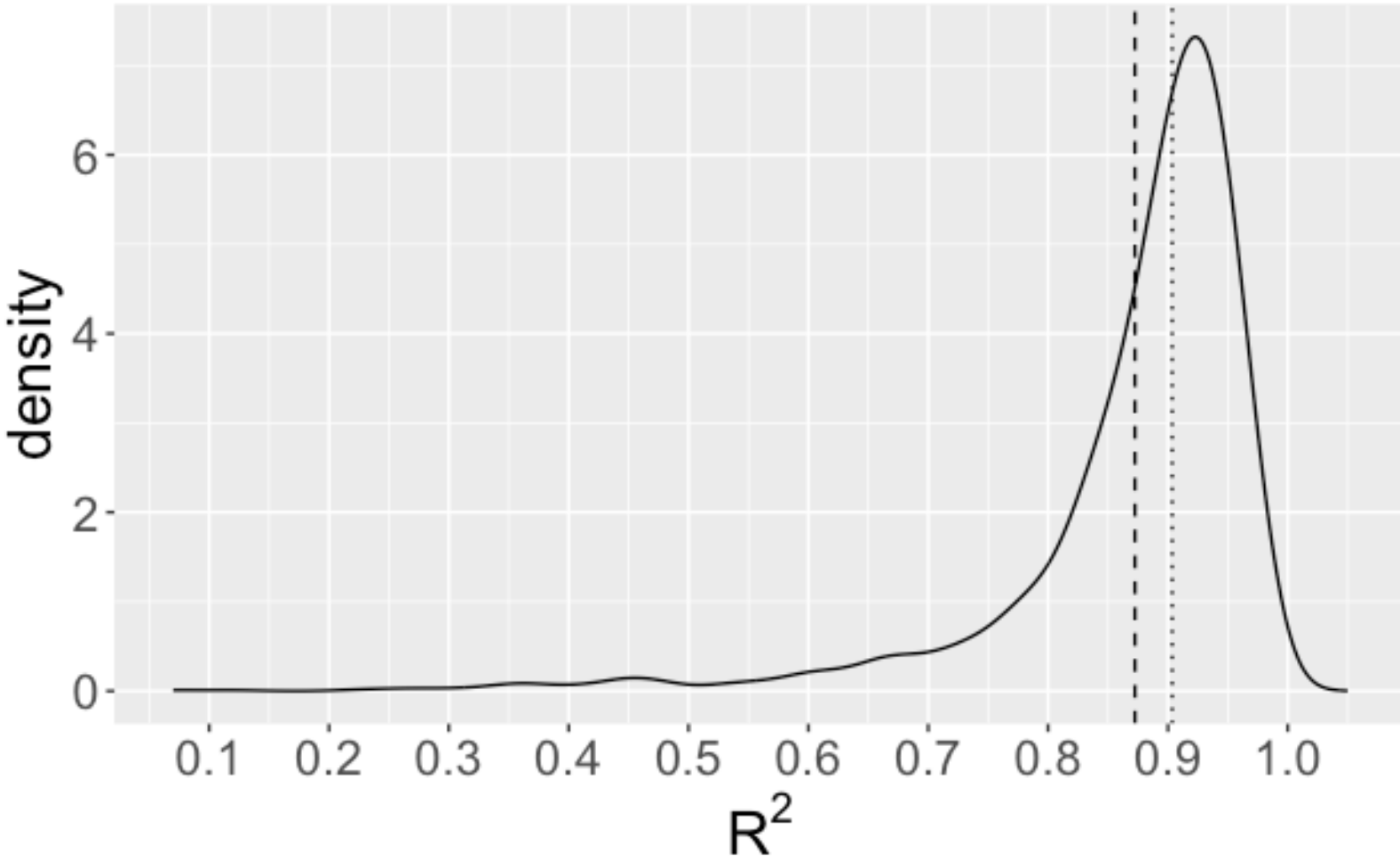

Figure 6: Distribution of $R^2$ values across 2398 fitted tokens. The dashed line indicates the mean (0.87); the dotted line indicates the median (0.90).

As seen in Figure 6, the fit of the model is generally excellent, with mean $R^2 = 0.87$ and median $R^2 = 0.90$. There are some outlier tokens with poorer fits, as seen in the leftward skew in the distribution. Of the 2398 analyzed tokens, 77 tokens have $R^2$ values below 0.6. 47 of these tokens have multiple velocity peaks, i.e., acceleration crosses zero multiple times. The other 30 tokens have additional bends in the velocity trajectory; i.e., acceleration approaches zero multiple times. Figure 7 displays example Hooke planes (acceleration by state) for a well-fit token ($R^2 = 0.97$; left) and a poorly fit token ($R^2 = 0.43$; right). It can be seen that the poorly fit token (Figure 7: right) has multiple zero-crossings in the acceleration curve, while the corresponding model-generated curve has only a single zero-crossing. In Section 5 we discuss potential sources of abnormal trajectories and the failure of the model to fit them well.

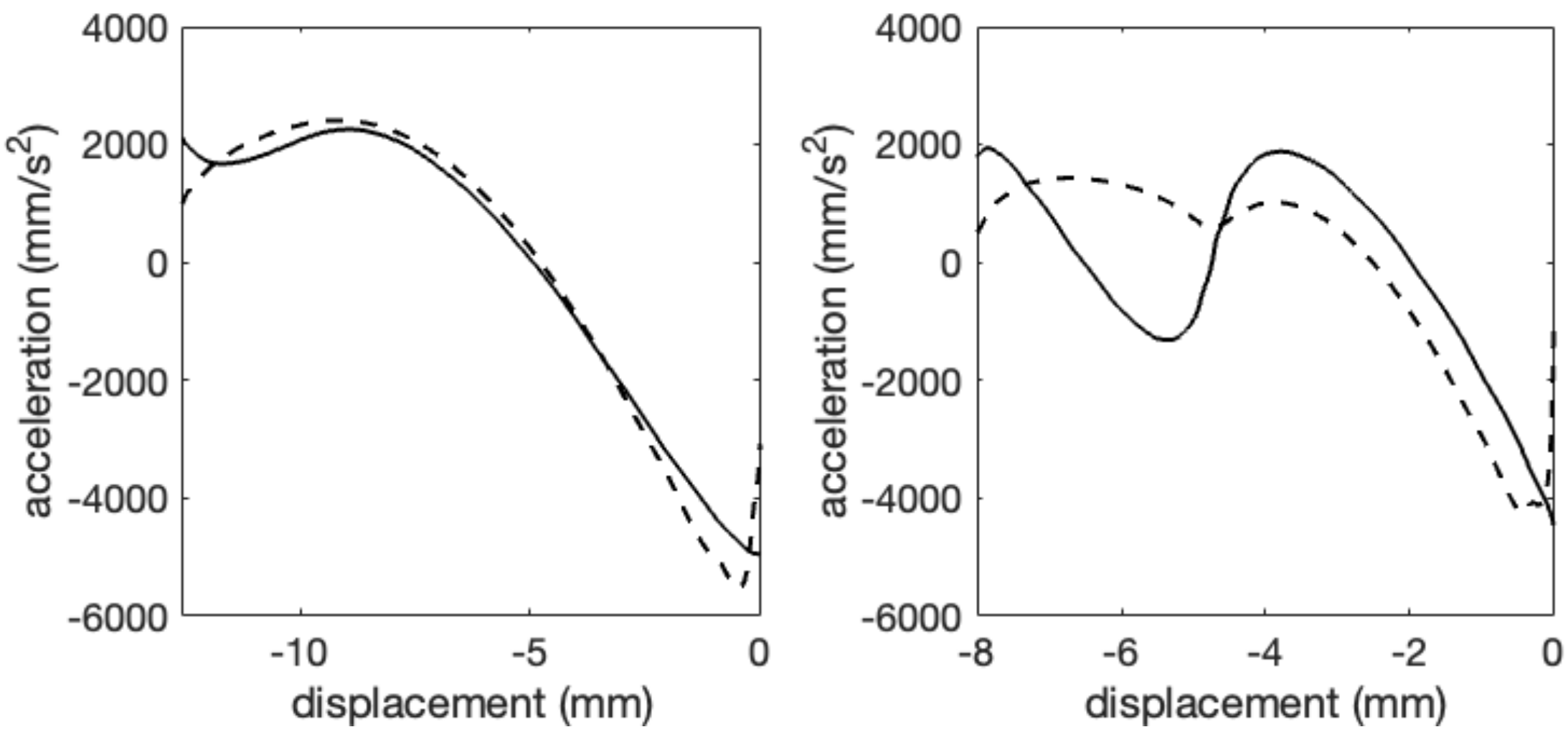

Figure 7: Hooke planes (acceleration by state) for a well-fit token ($R^2 = 0.97$; left) and a poorly fit token ($R^2 = 0.43$; right). Observed acceleration curves are solid; model-generated curves are dashed.

Figure 8 displays the distributions of the values of the parameters $r$ and $T$ fit to each trajectory.

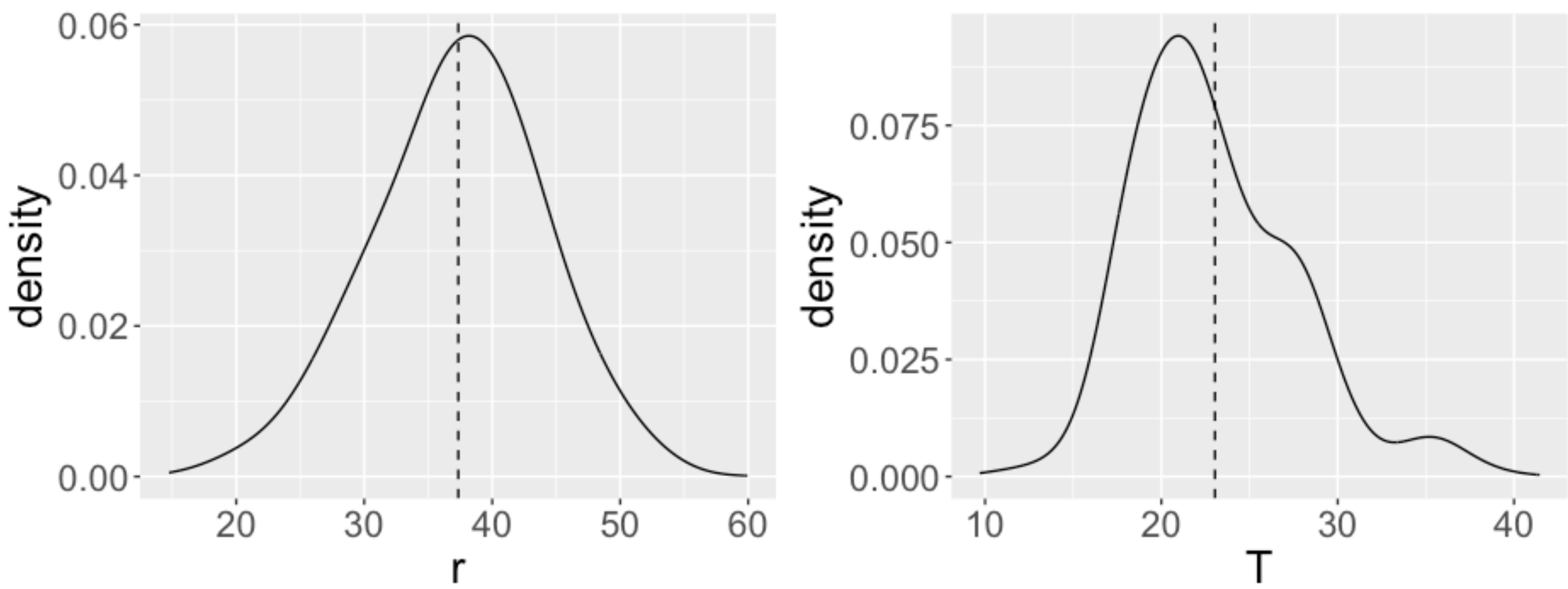

Figure 8: Distributions of fitted values for the parameters $r$ (left; $M = 37.35$, $SD = 6.73$) and $T$ (right; $M = 22.04$, $SD = 4.64$).

The fitted values of $r$ correlate strongly with the slopes of the regression lines fit to $\ln(\lambda)$ ($\rho = -.96$, $p < .001$; Figure 9: left), and the fitted values of $T$ correlate strongly with LA state at the timepoint of minimum velocity ($\rho = 1$, $p < .001$; Figure 9: right).

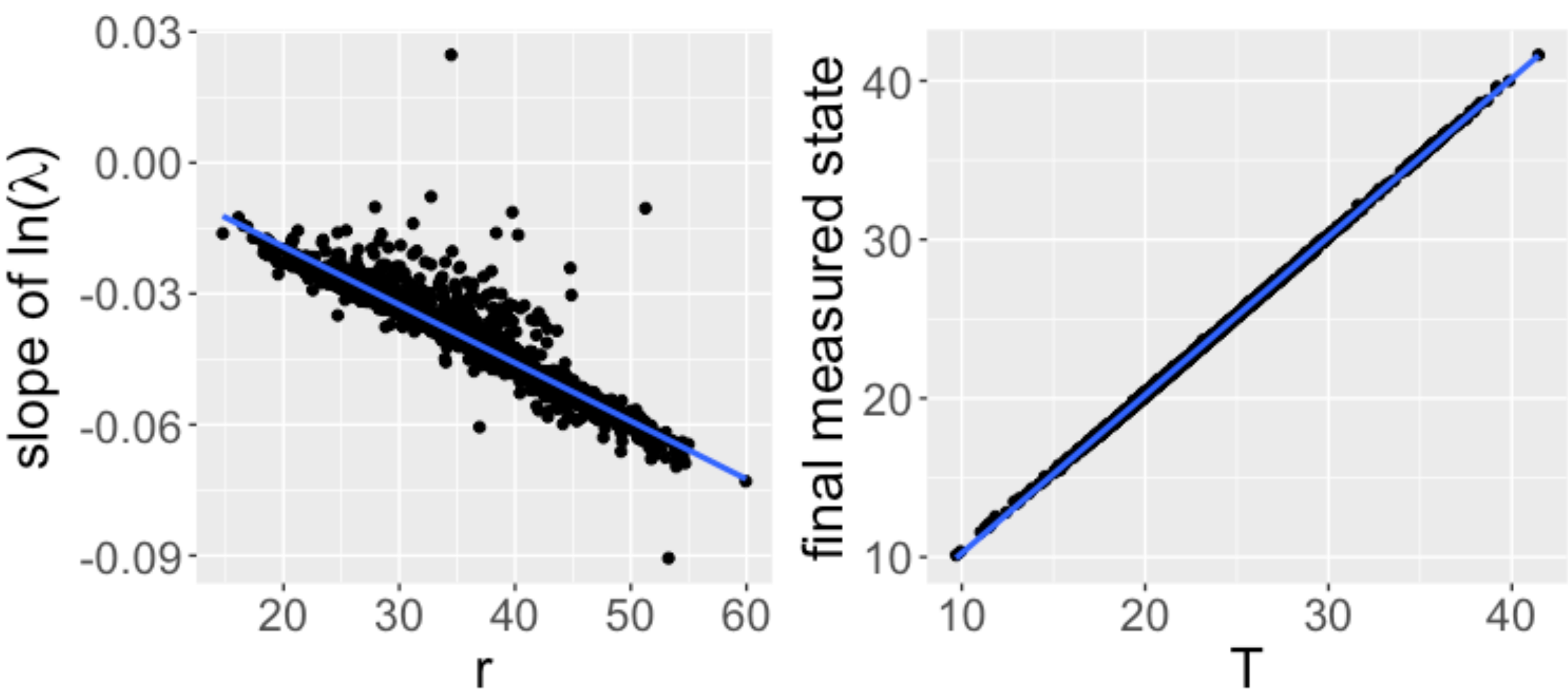

Figure 9: Relationship between fitted parameter values ($x$-axes) and corresponding measured values ($y$-axes).

Finally, consistent with our interpretation of $r$ as controlling movement rapidity, Figure 10 shows strong correlations between $r$ and movement duration ($\rho = -.88$, $p < .001$; Figure 10: left), and between $r$ and kinematic stiffness ($\rho = .91$, $p < .001$; Figure 10: right). Figure 10 is analogous to Figure 5, but with the model control parameter $r$ on the $x$-axis instead of its corresponding measured value, the slope of $\ln(\lambda)$.

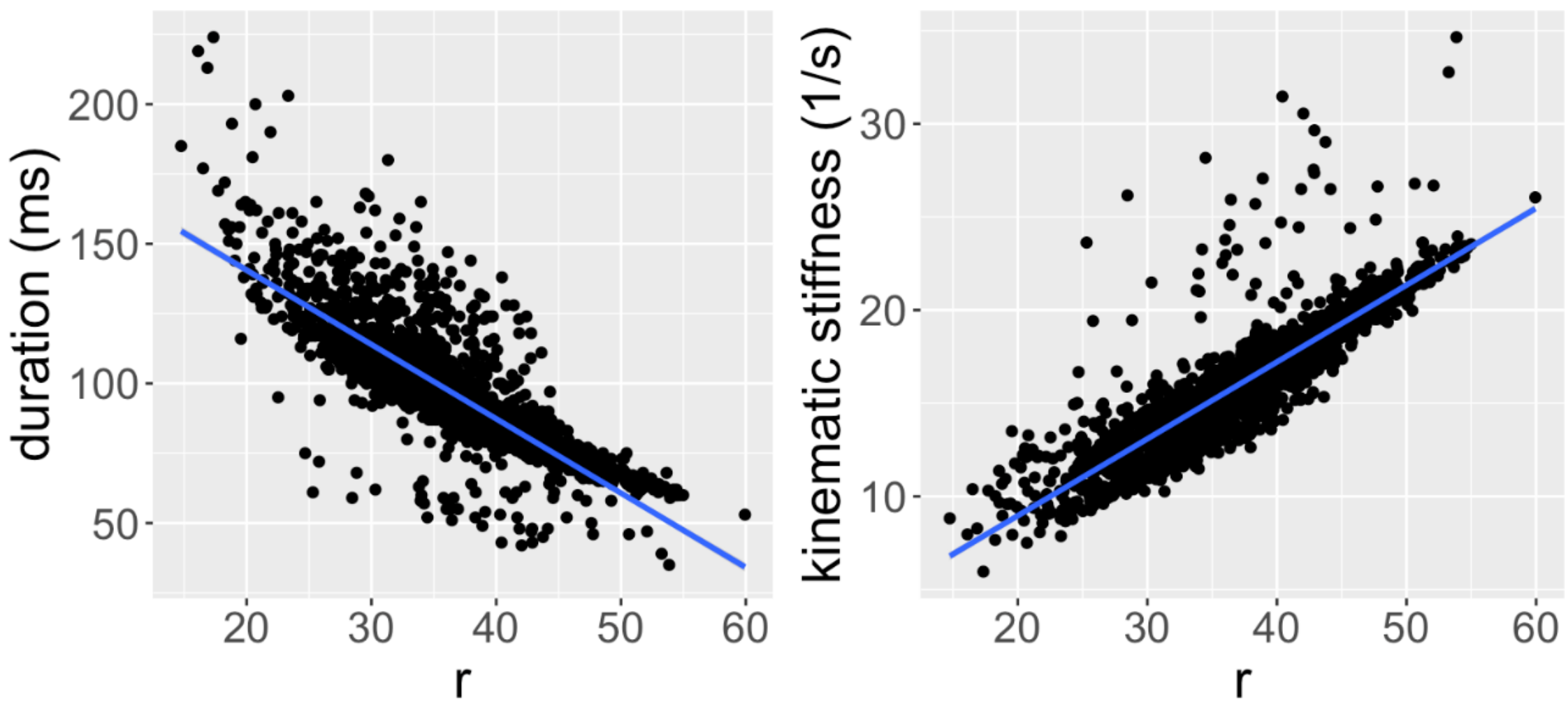

Figure 10: Relationship between the model control parameter $r$ ($x$-axes) and two kinematic variables: movement duration (left) and kinematic stiffness (right), analogous to Figure 5.

*4.3 Model simulation*

In Section 4.2, the entire observed state and velocity trajectories were used to fit the acceleration curve. Another way to evaluate the model is to *generate* trajectories from Eq. 5 using the fitted values of the control parameters $r$ and $T$. Then, generated trajectories can be compared to real trajectories. This is the approach we take in this section. This approach differs from that in Section 4.2 since we now consider only the *initial* state and *initial* velocity of the observed trajectories, along with the fitted parameters $r$ and $T$; the full trajectories of state, velocity, and acceleration are then generated solely from the model in Eq. 5. This demonstrates the utility of Eq. 5 as a generative model of speech articulation.

For each token, we set $x_0$ and $\dot{x}_0$ as their observed values, and calculated $\ddot{x}_0$ from Eq. 5 using the fitted values of $r$ and $T$. Next, $\dot{x}_1$ was calculated as $\dot{x}_0 + \ddot{x}_0 * dt$ (set to .001), $x_1$ was calculated as $x_0 + \dot{x}_0 * dt$, and $\ddot{x}_1$ was calculated from $\dot{x}_1$, $x_1$, $r$ and $T$ using Eq. 5. We iterated this process $t$ times until $\dot{x}_t$ fell below $\dot{x}_0$ (20% of measured peak velocity). Average simulated trajectories are displayed in Figure 11 (top) with average observed trajectories reproduced for comparison in Figure 11 (bottom).

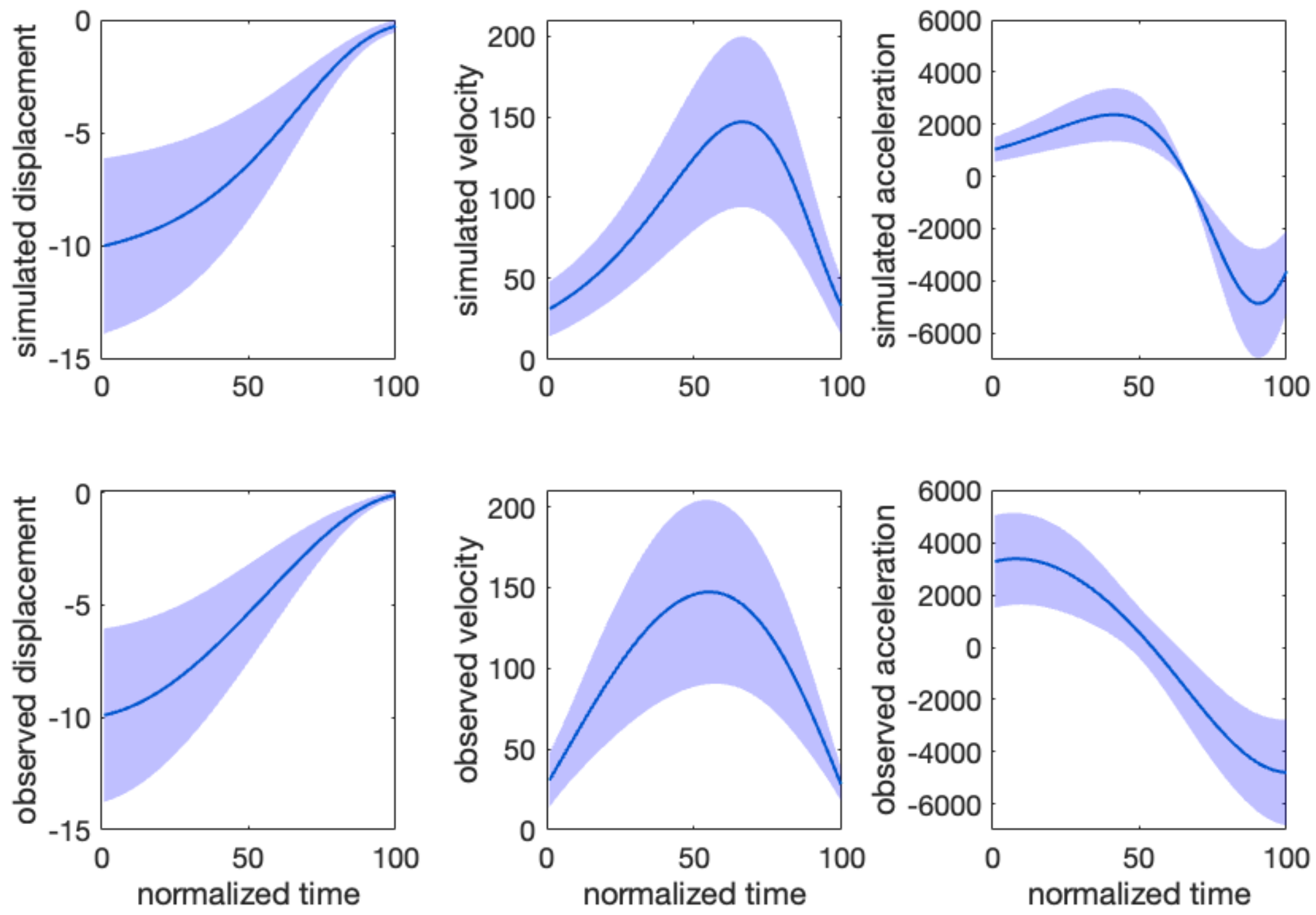

Figure 11: Average trajectories simulated from model fits (top) and observed in the data (bottom).

We calculated $R^2$ for the fit of each model-generated trajectory ($\ddot{x}$, $\dot{x}$, and $x$) to each corresponding observed trajectory. Overall, model-generated state trajectories $x$ were closest to the observed trajectories (mean $R^2 = 0.90$), followed by acceleration $\ddot{x}$ (mean $R^2 = 0.73$), followed by velocity $\dot{x}$ (mean $R^2 = 0.51$).

In order to further assess the relationship between model-generated and observed trajectories, we measured the same set of kinematic variables as those measured for the observed trajectories and displayed in Figure 3. Figure 12 displays the model-generated distributions.

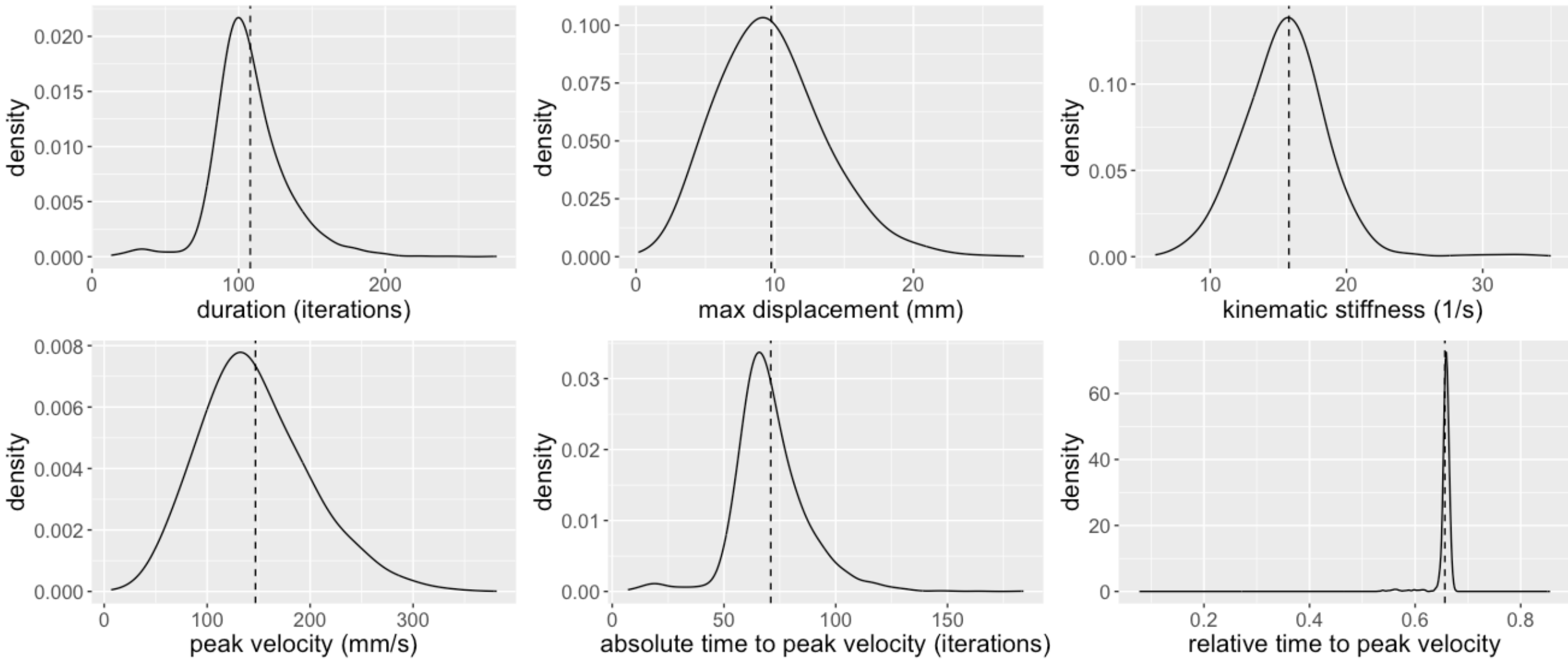

Figure 12: Distributions of kinematic variables measured from simulated trajectories. $x$-axis limits match those from Figure 2 for comparison. 18 tokens with extremely large kinematic stiffness values (stiffness $> M + 4 * SD$) were removed from this plot (total $n = 2380$).

Next, Figure 13 displays the relationships between the model-simulated kinematic variables and the corresponding observed variables.

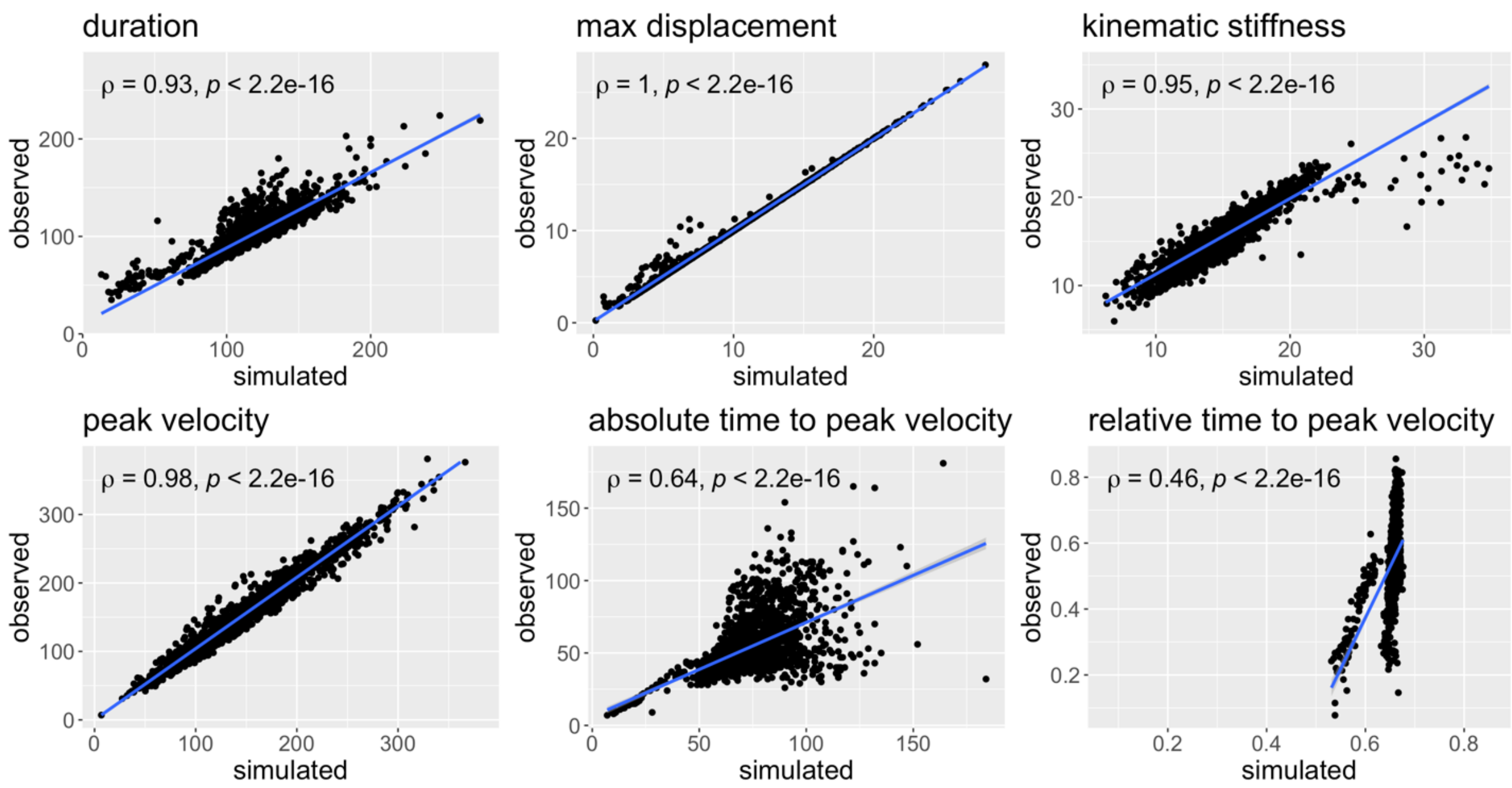

Figure 13: Relationships between model-simulated and observed kinematic variables. The same tokens that were removed from Figure 12 were also removed from these plots.

In general, simulated kinematic variables are highly correlated with observed kinematic variables (all $p$-values < .001). The weakest correlation was observed in relative time to peak velocity. Simulated trajectories exhibited very low variation in this variable (*SD* = 0.02) relative to the observed variation (*SD* = 0.09). We also note that there appear to be two distinct groups of data in the plot of relative time to peak velocity. On the suggestion of an anonymous reviewer, we investigated whether this distinction is related to duration. Figure 14 plots relative time to peak velocity by duration in the simulated data (left) and empirical data (right). For simulated movements with short duration—less than approximately 75 ms—relative time to peak velocity is overall earlier, more variable, and more strongly correlated with duration. For simulated movements with longer duration, relative time to peak velocity is overall later, less variable, and less correlated with duration. There is some overlap between the two groups, at durations from about 75 to 100 ms, which suggests that other factors besides duration play a role in distinguishing the two groups of data. Aspects of these patterns are visible in the empirical data as well (Figure 14: right). For example, at shorter durations—less than approximately 75 ms—there is a stronger correlation between duration and relative time to peak velocity than at longer durations. There are also some differences between the simulated and empirical data. The overall range of relative time to peak velocity is much greater in the empirical data. Additionally, the empirical data shows increased variability in relative time to peak velocity at longer durations. However, both the simulated data and the empirical data have ranges of duration values with low variability in relative time to peak velocity. In the empirical data, low variability is found at durations of around 80 ms. In the simulated data, the range of low variability is around 100 ms. As far as we know, the nonlinearity in the relation between movement duration and relative time to peak velocity found in our data has not been reported before. Our model captures aspects of this pattern but predicts an even sharper nonlinearity than we observe. We discuss possible reasons for mismatches between simulated and empirical data in Section 5.

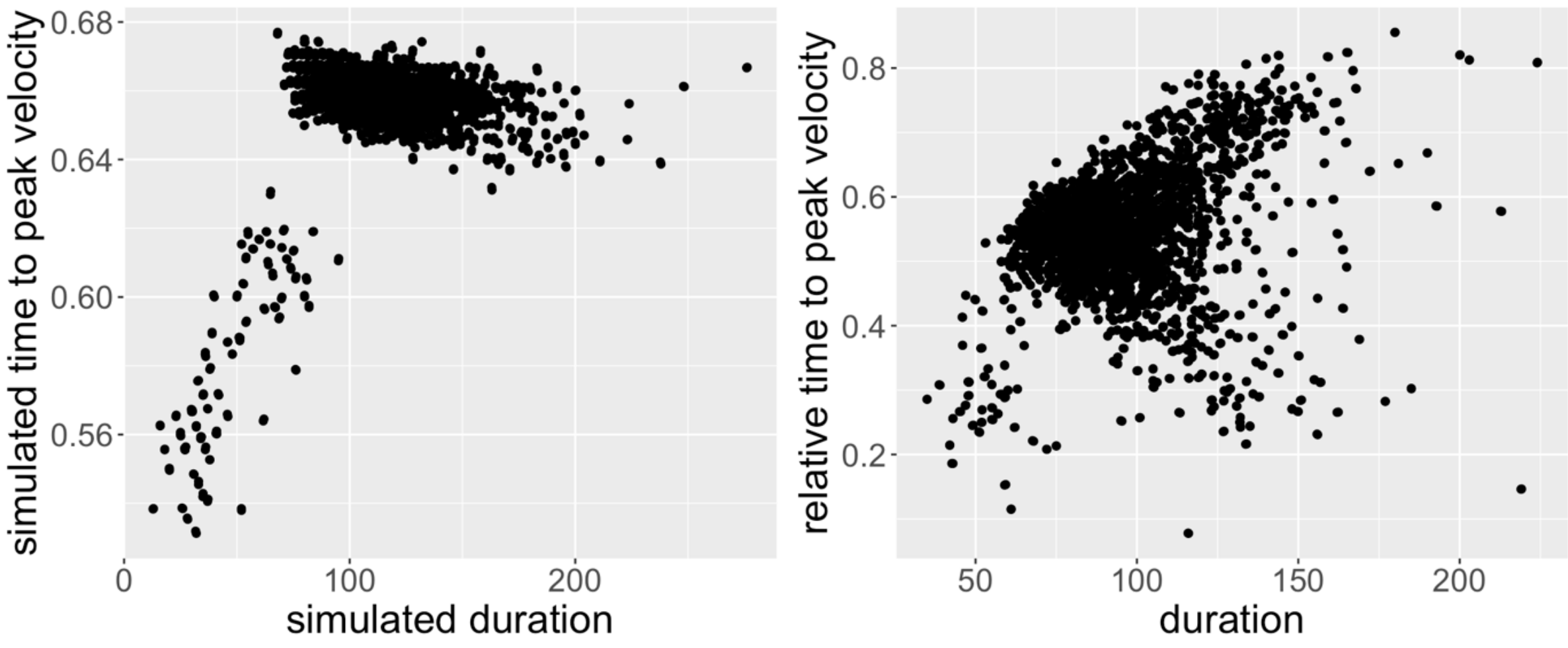

Figure 14: Relationship between duration and relative time to peak velocity in simulated (left) and empirical (right) data.

Finally, as seen in Figure 15 (left), model-simulated trajectories replicate the slightly nonlinear relationship between maximum displacement and peak velocity observed in both previous datasets (e.g., Ostry & Munhall, 1985) as well as the current dataset (Figure 15: right). It can be seen in this dataset, similar to previous datasets, that while the relationship appears linear for most of the observed range, it appears nonlinear at extreme values. Nonlinearity in the relationship is supported by the fact that a quadratic model of peak velocity by max displacement significantly improved over a linear model for observed ($F(1, 2396) = 51.59, p < .001$) and simulated ($F(1, 2396) = 24.44, p < .001$) trajectories. Recall that this nonlinearity was one motivation for the addition of a cubic term to the damped mass-spring model (Sorensen & Gafos, 2016). Our model captures this fact without additional control parameters.

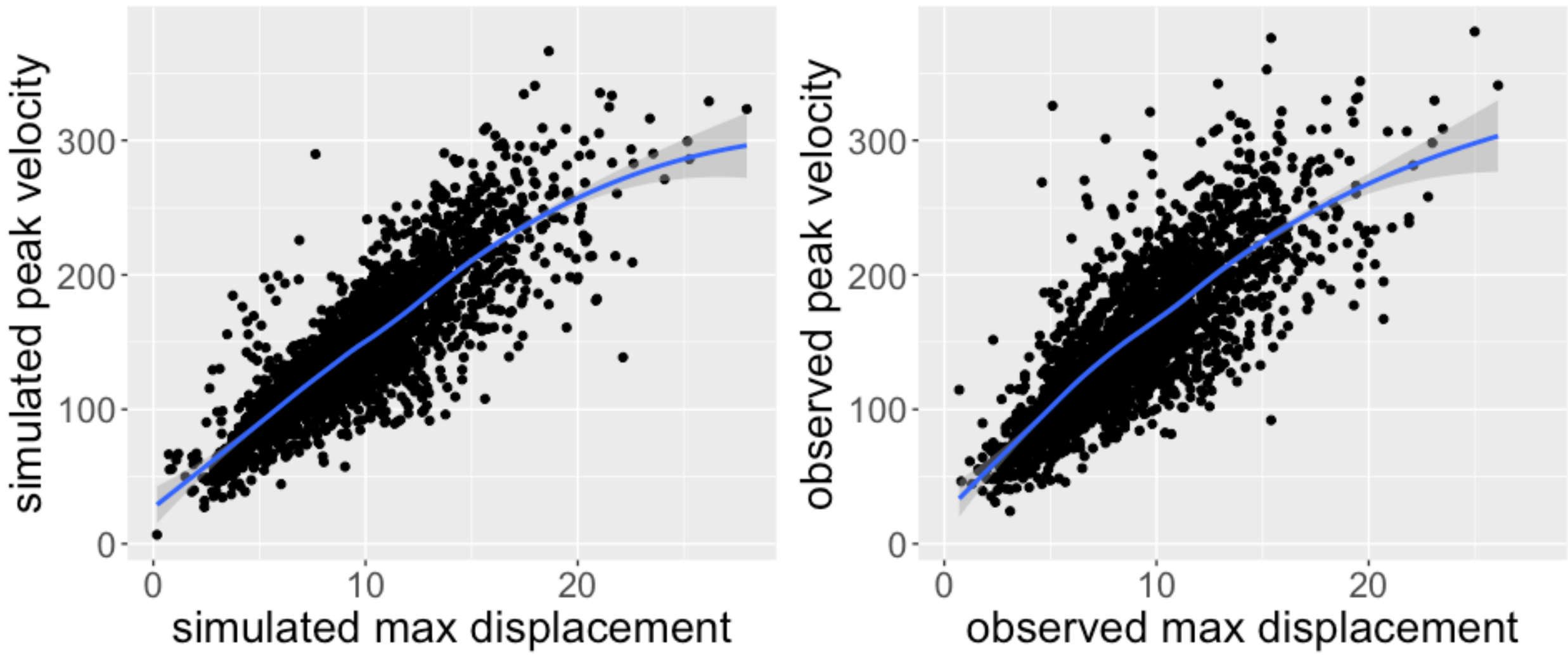

Figure 15: Relationship between maximum displacement and peak velocity in the simulated data (left) and the observed data (right). The line represents the fit of a LOESS regression.

## 5. Discussion

### *5.1 Interpretation of control parameters*

The model proposed in Section 4 provides an excellent fit to observed lip constriction trajectories (Figure 6; although some individual tokens were poorly fit), generates trajectories which are qualitatively similar to observed trajectories (Figure 11; although mean $R^2$ for velocity trajectories was only 0.51), and captures key kinematic properties of the trajectories (Figures 13, 15; although relative time to peak velocity had a correlation of only $\rho = .46$). The model also captures previously unreported aspects of the relationship between duration and relative time to peak velocity (Figure 14; although variability in relative time to peak velocity was overall much greater in the empirical data relative to the simulated data). The model achieves these empirical successes despite having only two control parameters: $T$, controlling the position of the target (Figure 9), and $r$, controlling the rapidity of the movement (Figure 10). The model thus has fewer parameters than the damped mass-spring model, as well as related models using time-dependent activation (Byrd & Saltzman, 1998; Kröger et al., 1995) and cubic nonlinearity (Sorensen & Gafos, 2016). Fewer parameters means that each parameter is more clearly interpretable. With respect to general movement control, analogous parameters (in particular, spatial direction and speed) have been found to be encoded in neural populations in monkey motor cortex during arm reaching movements, consistent with the notion that $T$ and $r$ are primary dimensions of movement control (Moran & Schwartz, 1999). In addition, $T$ and $r$ have clear links to linguistic dimensions. For the lip aperture (LA) trajectories examined here, $T$ corresponds to the phonological dimension of constriction degree (CD: Browman & Goldstein, 1989) or manner of articulation, which distinguishes stops (small CD), fricatives (intermediate CD), and approximants (large CD). For other kinds of movements, e.g., lingual movements, $T$ may also correspond to constriction location (CL) or place of articulation. Conceptualizing $T$ as a parameter of phonological control offers a way to concretely implement different theoretical approaches to phonological targets. In the present study, we held $T$ constant at a single value for the duration of each movement. This is consistent with the notion that phonological targets are stable and invariant (e.g., Browman & Goldstein, 1989), in contrast to the time-varying articulatory trajectories which they contribute to. However, it would also be straightforward to select each $T$ value from a "window" (Keating, 1990) or a "convex region", as in the DIVA model (Guenther, 1995), or from a distribution of probability or activation, as in exemplar theory (Pierrehumbert, 2001) or Dynamic Field Theory (DFT: Roon & Gafos, 2016; Schöner et al., 2016). In particular, if $T$ is selected from a neural field, as in DFT, this would allow $T$ to vary over the course of a single movement, since neural fields are themselves dynamic.

The other control parameter, $r$, corresponds to the speed or rapidity of movement (see Figure 10), roughly analogous to stiffness $k$ in the damped mass-spring model (see Section 1.1). Articulatory timing is a fundamental dimension of phonological control across levels of description, i.e., gestural, segmental, and suprasegmental (e.g., Shaw, 2022). Across languages, articulatory movements are temporally structured, composing into coordinative units like segments and syllables. In particular, there is evidence that movement speed (corresponding to the control parameter $r$) is manipulated to control the degree of coarticulation between overlapping movements (Du et al., 2023, 2025; Du & Gafos, 2023; Roon et al., 2021; Shaw & Chen, 2019), which has been argued to be language-specific and thus part of the learned phonological system (Shaw, 2022). In some languages, temporal properties of individual articulatory movements can signal lexical information through consonant gemination and vowel length (e.g., Burroni et al., 2025). Temporal properties of articulatory movements are also utilized to express phrase-level prosodic information. For example, movements are generally slower at phrase boundaries (Byrd & Saltzman, 1998, 2003; Cho, 2016) and they are longer in duration when part of informationally prominent words (Katsika & Tsai, 2021; Liu et al., 2023; Roessig et al., 2019; Roessig & Mücke, 2019). Along the lines of the discussion above regarding the parameter $T$, conceptualizing $r$ as a dimension of phonological control has the potential to shed light on questions about the nature of articulatory timing. Are timing goals single stable values, or are they more flexible? A promising possibility is that $r$ is generated from a dynamic neural field under the simultaneous influence of multiple linguistic forces. Under this possibility, phonological representations (as field input distributions) are stable, while the $r$ value for any given movement is the product of multiple input distributions—relating to phonology, prosody, and inter-movement coordination—interacting in the same neural field under the additional influence of stochastic noise (e.g., Shaw & Tang, 2023). In this way, $r$ can simultaneously map to multiple kinds of phonological primitives—e.g., intrinsic segmental duration, suprasegmental coordination, context-dependent prosody—each modeled as a separate input to the same neural field controlling $r$. Fleshing out such a neural field model of articulatory timing control is beyond the scope of this paper, but would be a fruitful area for future work, given the potential of these kinds of models for unifying disparate empirical findings and theoretical insights, and for generating novel empirical predictions (Schöner et al., 2016; Stern, 2025).

We proposed that $T$ generally corresponds to the phonological dimensions of place and manner of articulation (or contriction location and constriction degree) while $r$ corresponds to phonological dimensions of gemination and length, as well as prosodic dimensions like phrase boundaries and informational prominence. However, it is not necessarily the case that the phonological dimensions indexed by each control parameter are so clearly separable. Another possibility is that a single phonological dimension can be indexed through an interaction of the two control parameters. For instance, linguistic prosody can modulate both spatial and temporal aspects of articulatory movements (Byrd & Saltzman, 1998, 2003; Katsika et al., 2014; Roessig et al., 2019; Roessig & Mücke, 2019). Under this possibility, we might expect systematic covariation between $r$ and $T$. As a first-pass investigation of this issue, Figure 16 displays the relationship between the fitted $r$ and $T$ values for each token.

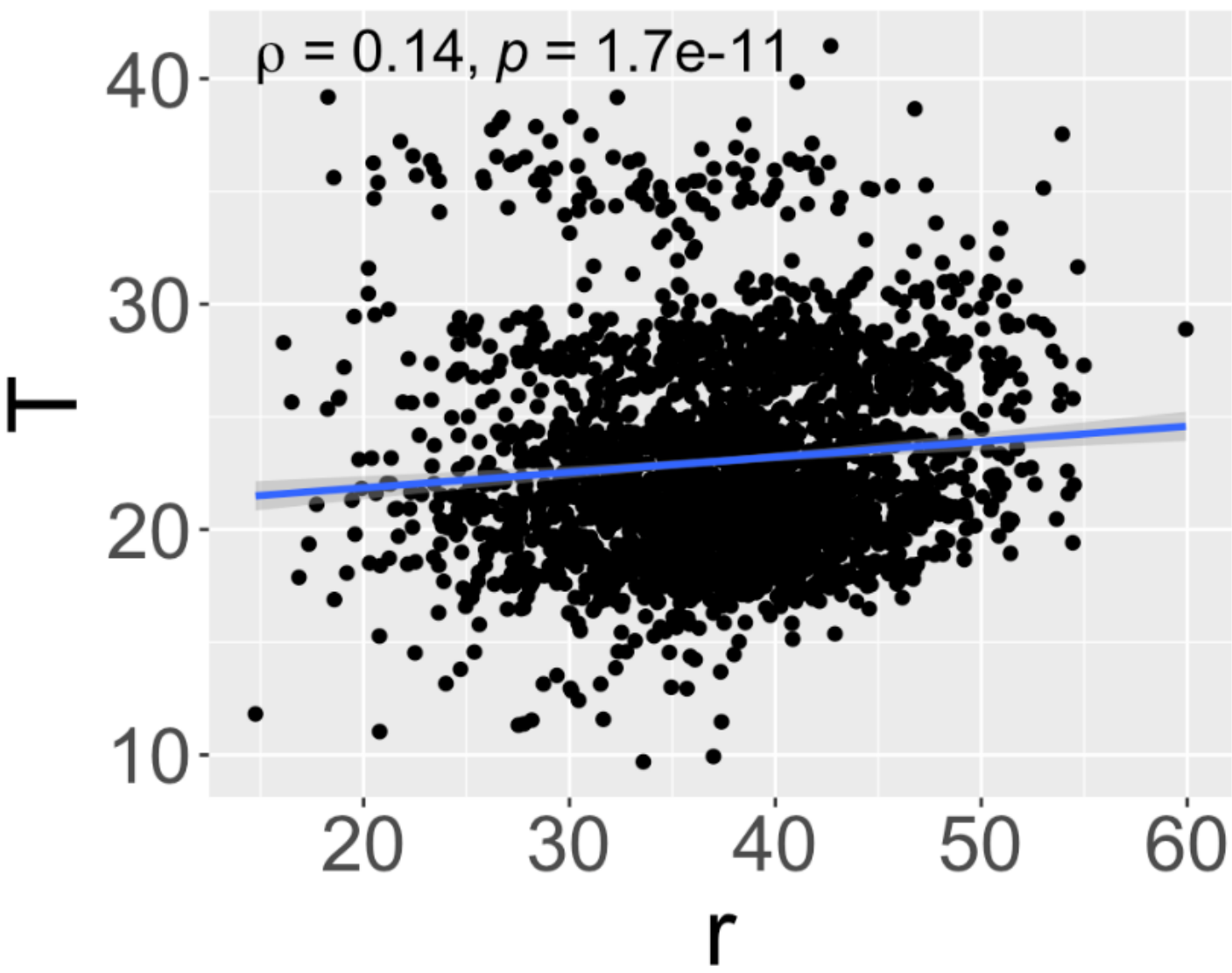

Figure 16: Relationship between fitted $r$ and $T$ values for each token.

There is a weak but significant positive correlation between $r$ and $T$ ($\rho = .14$, $p < .001$). In other words, movements with more extreme spatial targets (smaller $T$) also tend to have smaller $r$. Note that, for any value of $r$, the model generates faster movements for more extreme spatial displacements (see, e.g., Figure 15). The small positive correlation between $r$ and $T$ can be seen as mitigating this relationship: given an extreme spatial target, the control system attenuates speed (smaller $r$) relative to the speed that would have been produced under an unchanged $r$ value. From the perspective of the damped mass-spring model, an analogous observation would be a correlation between $k$ and $T$. To our knowledge, the correlation between these parameters has not been explored in past work. If $r$ and $T$ are governed by dynamic neural fields, as discussed above, then this correlation might arise from coupling between fields. Work on lexical semantic processing in the DFT framework has modeled dependencies between conceptual dimensions as patterned coupling between neural fields (Stern & Piñango, 2024). The result in Figure 16 suggests that neural field coupling may also apply to dimensions of articulatory control. This proposal is speculative at this point. The relationship between control parameters remains a fruitful area for future work.

### *5.2 Model characteristics*

Like the family of damped mass-spring models, the system in Eq. 5 is second-order, referencing the second derivative $\ddot{x}$ or acceleration. It is noteworthy that, although our starting point was the first-order equation in Eq. 2, formalizing the observed temporal variation in $\lambda$ led to the second-order equation in Eq. 5. It is not surprising that we arrived at a second-order description, given that the empirical shapes of velocity curves are difficult to capture with first-order dynamics, as described in Section 1. Like the original version of the damped mass-spring model (Ostry & Munhall, 1985) as well as the version with a cubic term (Sorensen & Gafos, 2016) but unlike the version with time-dependent activation (Byrd & Saltzman, 1998; Kröger et al., 1995), our model is autonomous: the control law only references the internal state of the system $x$ and $\dot{x}$, without referencing an external time variable $t$. Moreover, our model is nonlinear. Nonlinear dynamical systems are generally more difficult to analyze than linear systems. Some nonlinear systems are *chaotic*, exhibiting sensitive dependence on initial conditions and unpredictable behavior. Thus, although our model is simple in that it has only two control parameters, each with a clear linguistically relevant interpretation, the feature of nonlinearity can be seen as a source of complexity in the model. For the range of parameter values and initial conditions examined in this paper, simulated trajectories closely matched observed trajectories. However, in future work, we plan to probe the behavior of the model under a wider range of conditions in order to determine whether the model is susceptible to chaotic behavior.

The model fits (mean $R^2 = 0.87$) are comparable to or slightly higher than reported fits of the damped mass-spring model to data (Kuberski & Gafos, 2023). However, it is difficult to directly compare the $R^2$ values from this study to those from Kuberski & Gafos (2023). First, the details of the fitting procedures differ: Kuberski & Gafos (2023) fixed the target parameter $T$ at the final observed value (and the parameter $m$ at 1), and then fit the parameters $k$ and $b$, both of which affect the shape of the trajectory but not the position of the point attractor.[5] Moreover, the data analyzed in Kuberski & Gafos (2023) came from a syllable repetition task, different from the sentence production task in the present dataset. Finally, Kuberski & Gafos (2023) examined tongue trajectories, which are more complex in that they theoretically involve multiple task dimensions, unlike the lip aperture (LA) trajectories examined in this study. In future work, we plan to address the question of the task dimensionality of articulatory movements following the approach reported here.

### *5.3 The role of single movement dynamics in speech articulation*

Our model (Eq. 5) captures the kinematics of a single task variable during a single movement. In this paper, we focused on the lip aperture variable during a constriction movement for a consonant, but we intend our approach to be extendable to other task variables (e.g., lingual variables) and other kinds of movements (e.g., releases), as we will discuss in Section 5.4. However, speech articulation is far more complex than any single movement. In natural speech, movements are produced together in coordinated ensembles. Several models have been proposed for sequencing articulatory movements, i.e., triggering their onsets, based on a coupled oscillator representation (Nam & Saltzman, 2003) or competitive queuing (Bohland et al., 2010; Tilsen, 2016). There is also evidence that movements continue to be influenced by other movements after becoming active. In particular, there is evidence that the relative timing of movement *targets*, in addition to movement onsets, is controlled (Kramer et al., 2023; Perkell et al., 1993; Shaw & Chen, 2019; for discussion, see Iskarous & Pouplier, 2022; Turk & Shattuck-Hufnagel, 2020). One way to capture such online coordination in a dynamical framework would be to couple the dynamics of individual movements, such that each movement depends not only on its own state but also the state of other coordinated movements, throughout its duration. Another aspect of speech articulation which is beyond single movement dynamics, per se, is parameter selection. The control parameters which shape articulatory kinematics are themselves the result of neural activity (e.g., Chartier et al., 2018), which has its own dynamics (see Section 5.1). Some progress has been made in explicitly coupling articulatory dynamics with the neural dynamics of parameter selection in the DFT framework (Chaturvedi & Shaw, 2025; Kim & Tilsen, 2025; Kirkham & Strycharczuk, 2024; Shaw, 2025; Tilsen, 2019). These additional aspects of speech articulation—sequencing, coordination, and parameter selection—leave their own imprints on single movement kinematics which cannot be captured by our model, since it treats individual movements as isolated. This is not to mention noise, which exists in each of these components and also leaves its imprint on kinematics (e.g., Mücke et al., 2024; Parrell et al., 2023). It is from this perspective that we now consider several apparent shortcomings of our model.

[5] A thorough comparison of the fits of different dynamical models of articulation (described in Sections 1.1 and 1.2) to the same dataset using the same fitting methods would be valuable, but is beyond the scope of this paper. In particular, fitting the nonlinear second-order model of Sorensen & Gafos (2016) requires imposing scaling laws on the control parameters, or scaling spatial displacement trajectories, in order to avoid model instability at displacement values greater than one (Kirkham, 2025b).

While model fits are overall very good, there are a number of outlying tokens which were poorly fit, even after excluding non-monotonic trajectories ($n$ = 87, 2.8%) where LA was moving away from its final state for at least one sample. Moreover, in comparing simulated trajectories to observed trajectories, the average fit of the velocity curves was relatively poor (mean $R^2 = 0.51$). Figure 17 displays the observed (solid line) and simulated (dashed line) displacement, velocity, and acceleration trajectories of a representative poorly fit token.

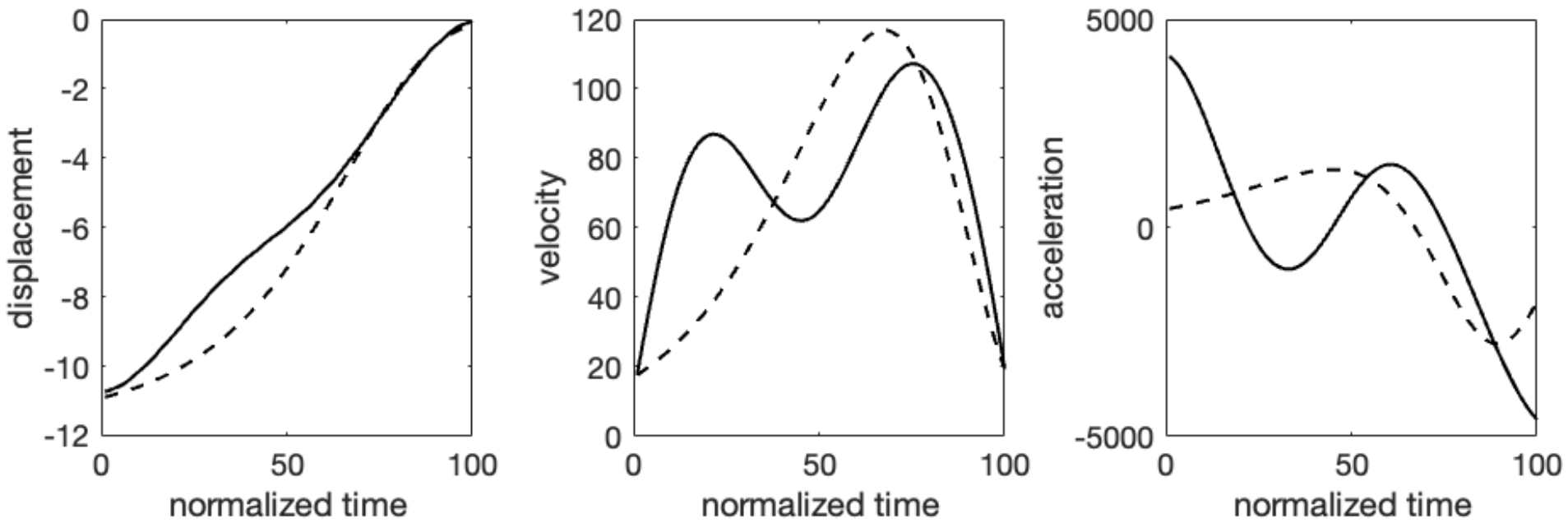

Figure 17: Observed (solid line) and simulated (dashed line) displacement (left), velocity (center), and acceleration (right) trajectories of a representative poorly fit token.

It can be seen that the magnitude of observed acceleration is initially very large, with a correspondingly steep slope in velocity. This initial rapidity then attenuates: the magnitude of velocity decreases before increasing again to another peak, and then finally decreasing again as the target is approached. In the displacement curve, this jerkiness is not particularly salient, and as a result, the model is able to match the displacement curve quite well ($R^2 = 0.92$). However, the model-generated acceleration and velocity curves are much smoother than the observed curves, leading to a worse fit for acceleration ($R^2 = 0.38$) and an extremely poor fit for velocity ($R^2 = –0.55$). This is largely because the model is unable to generate trajectories with multiple velocity peaks. Relatedly, model-generated velocity trajectories show very little variation in the relative timing of the single velocity peak as a ratio of total movement duration, particularly for longer duration (> 75 ms) movements (see Figures 13 and 14, and related discussion). While variation in the two parameters $r$ and $T$ is able to capture some of the observed variation in relative time to peak velocity ($\rho = .46$, $p < .001$), the correlation is relatively small compared to the other kinematic variables examined. It is a feature of the model that relative time to peak velocity is very stable under variation in model parameter values, more stable than in real articulatory movements.

One way to address these apparent shortcomings of our model would be to complexify the dynamics in order to more closely fit the data. In general, more variation in the shapes of trajectories can be captured by adding more control parameters to the dynamics (e.g., Brunton et al., 2016; Kirkham, 2025a). This is essentially the strategy pursued in the work described in Section 1. Byrd & Saltzman (1998) captured variation in relative time to peak velocity induced by the presence of a prosodic boundary by varying the trajectory of the time-dependent activation term $a(t)$. In the model in Sorensen & Gafos (2016), some variation in relative time to peak velocity can be controlled by the $d$ parameter weighting the nonlinear term. As discussed in Section 1, there is a tradeoff between data fitting and parameter interpretability. Each of the two control parameters of our model has a clear interpretation with respect to both the neural dynamics of movement control (Moran & Schwartz, 1999) and dimensions of linguistic information, as discussed in Section 5.1. While more kinematic variation could be captured by adding more terms to our model, we are inclined to preserve the interpretability of our model parameters. This leads us to hypothesize that the kinematic variation not captured by our model arises from factors outside of individual movement dynamics. Some of these factors, as discussed above, are sequencing, coordination, parameter selection, and noise. We hope that our proposal for individual movement dynamics can offer a fruitful starting point to test this hypothesis.

### *5.4 Beyond labial constrictions*

As a final point, the data motivating our model comes only from labial constriction movements, which is a small subset of the movement types relevant for speech articulation. There is some evidence that movements of different articulators have similar kinematic properties. For example, both laryngeal adduction and lingual constriction movements show a robust correlation between movement amplitude and peak velocity (Munhall et al., 1985). This pattern has motivated a common task dynamics with the same set of control parameters regardless of the articulator(s) involved in achieving the task (Saltzman & Munhall, 1989). Besides offering a new dynamical model, we have presented a method for discovering dynamics from data, which can be applied to other types of speech articulatory movements. A key assumption in our approach is that articulatory movements are defined by point attractor dynamics. This departs from other work, inspired by rhythmic (non-speech) movements, that articulation follows oscillatory dynamics (e.g., Cooke, 1980). From the starting assumption of point attractor dynamics, we empirically evaluated the relationship between instantaneous velocity and distance to target over time. This same procedure could be repeated with other types of speech movements to evaluate whether we would find converging dynamics.

## 6. Conclusion

We discovered that the ratio of instantaneous displacement, $-x + x_f$, to instantaneous velocity, $\dot{x}$, decays exponentially over the course of labial constriction movements for /b/ and /m/ in English and Mandarin in two different prosodic contexts. Formalizing this discovery led to the derivation of an autonomous, nonlinear, second-order dynamical model of speech articulatory movements. The model achieves excellent fits to empirical trajectories with only two control parameters, $T$ and $r$. $T$ controls the position of the target, and $r$ controls the rapidity of the movement. These parameters, by hypothesis, are under cognitive control and constitute dimensions of phonological contrast in language. As a basic dynamical model of the control of individual articulatory movements, it offers a starting point from which to understand additional influences on movement, like prosody, inter-movement coordination, and stochastic noise in motor control.

## 7. References

Beckman, M. E., & Edwards, J. (1992). Intonational categories and the articulatory control of duration. In *Speech perception, production and linguistic structure*.

Bohland, J. W., Bullock, D., & Guenther, F. H. (2010). Neural representations and mechanisms for the performance of simple speech sequences. *Journal of Cognitive Neuroscience*, *22*(7), 1504–1529. https://doi.org/10.1162/jocn.2009.21306

Browman, C. P., & Goldstein, L. (1989). Articulatory gestures as phonological units. *Phonology*, *6*(2), 201–251. https://doi.org/10.1017/S0952675700001019

Brunton, S. L., Proctor, J. L., & Kutz, J. N. (2016). Discovering governing equations from data by sparse identification of nonlinear dynamical systems. *Proceedings of the National Academy of Sciences*, *113*(15), 3932–3937. https://doi.org/10.1073/pnas.1517384113

Burroni, F., Kawahara, S., & Shaw, J. A. (2025). Articulatory correlates of consonantal length contrasts: The case of Japanese mimetic geminates. *JASA Express Letters*, *5*(1), 015201. https://doi.org/10.1121/10.0034762

Byrd, D., & Saltzman, E. (1998). Intragestural dynamics of multiple prosodic boundaries. *Journal of Phonetics*, *26*(2), 173–199. https://doi.org/10.1006/jpho.1998.0071

Byrd, D., & Saltzman, E. (2003). The elastic phrase: Modeling the dynamics of boundary-adjacent lengthening. *Journal of Phonetics*, *31*(2), 149–180. https://doi.org/10.1016/S0095-4470(02)00085-2

Chartier, J., Anumanchipalli, G. K., Johnson, K., & Chang, E. F. (2018). Encoding of Articulatory Kinematic Trajectories in Human Speech Sensorimotor Cortex. *Neuron*, *98*(5), 1042-1054.e4. https://doi.org/10.1016/j.neuron.2018.04.031

Chaturvedi, M., & Shaw, J. A. (2025). A Dynamic Neural Model of Tonal Downstep. *Proceedings of the Annual Meetings on Phonology*, *1*(1). https://doi.org/10.7275/amphonology.3021

Cho, T. (2016). Prosodic Boundary Strengthening in the Phonetics–Prosody Interface. *Language and Linguistics Compass*, *10*(3), 120–141. https://doi.org/10.1111/lnc3.12178

Cooke, J. D. (1980). The Organization of Simple, Skilled Movements. In G. E. Stelmach & J. Requin (Eds.), *Advances in Psychology* (pp. 199–212). North-Holland. https://doi.org/10.1016/S0166-4115(08)61946-9

Cooper, W. E., Eady, S. J., & Mueller, P. R. (1985). Acoustical aspects of contrastive stress in question-answer contexts. *The Journal of the Acoustical Society of America*, *77*(6), 2142–2156. https://doi.org/10.1121/1.392372

Du, S., & Gafos, A. I. (2023). Articulatory overlap as a function of stiffness in German, English and Spanish word-initial stop-lateral clusters. *Laboratory Phonology*, *14*(1), Article 1. https://doi.org/10.16995/labphon.7965

Du, S., Kuberski, S., & Gafos, A. I. (2023). How measures of gestural overlap relate to dynamics: Evidence from German and English word-initial stop-lateral clusters. *Proceedings of the 20th International Congress of Phonetic Sciences (ICPhS)*, 2164–2168.

Du, S., Kuberski, S. R., & Gafos, A. I. (2025). Towards a dynamical account of inter-segmental coordination. *Journal of Phonetics*, *109*, 101392. https://doi.org/10.1016/j.wocn.2025.101392

Elie, B., Lee, D. N., & Turk, A. (2023). Modeling trajectories of human speech articulators using general Tau theory. *Speech Communication*, *151*, 24–38. https://doi.org/10.1016/j.specom.2023.04.004

Elie, B., Šimko, J., & Turk, A. (2024). Optimization-based planning of speech articulation using general Tau Theory. *Speech Communication*, *160*, 103083. https://doi.org/10.1016/j.specom.2024.103083

Fowler, C. A. (1980). Coarticulation and theories of extrinsic timing. *Journal of Phonetics*, *8*, 113–133.

Garcia, D. (2010). Robust smoothing of gridded data in one and higher dimensions with missing values. *Computational Statistics and Data Analysis*, *54*(4), 1167–1178. https://doi.org/10.1016/j.csda.2009.09.020

Guenther, F. H. (1995). Speech sound acquisition, coarticulation, and rate effects in a neural network model of speech production. *Psychological Review*, *102*(3), 594–621. https://doi.org/10.1037/0033-295X.102.3.594

Iskarous, K. (2017). The relation between the continuous and the discrete: A note on the first principles of speech dynamics. *Journal of Phonetics*, *64*, 8–20. https://doi.org/10.1016/j.wocn.2017.05.003

Iskarous, K., & Pouplier, M. (2022). Advancements of phonetics in the 21st century: A critical appraisal of time and space in Articulatory Phonology. *Journal of Phonetics*, *95*, 101195. https://doi.org/10.1016/j.wocn.2022.101195

Katsika, A., Krivokapić, J., Mooshammer, C., Tiede, M., & Goldstein, L. (2014). The coordination of boundary tones and its interaction with prominence. *Journal of Phonetics*, *44*(1), 62–82. https://doi.org/10.1016/j.wocn.2014.03.003

Katsika, A., & Tsai, K. (2021). The supralaryngeal articulation of stress and accent in Greek. *Journal of Phonetics*, *88*, 101085. https://doi.org/10.1016/j.wocn.2021.101085

Keating, P. A. (1990). The window model of coarticulation: Articulatory evidence. In M. E. Beckman & J. Kingston (Eds.), *Papers in Laboratory Phonology I: Between the Grammar and the Physics of Speech* (pp. 451–470). Cambridge University Press.

Kelso, J. A., Vatikiotis-Bateson, E., Saltzman, E. L., & Kay, B. (1985). A qualitative dynamic analysis of reiterant speech production: Phase portraits, kinematics, and dynamic modeling. *The Journal of the Acoustical Society of America*, *77*(1), 266–280. https://doi.org/10.1121/1.392268

Kim, S.-E., & Tilsen, S. (2025). The Gesture-Field-Register (GFR) framework for modeling F0 control. *Journal of Phonetics*, *110*, 101410. https://doi.org/10.1016/j.wocn.2025.101410

Kirkham, S. (2025a). Discovering Dynamical Laws for Speech Gestures. *Cognitive Science*, *49*(5), e70064. https://doi.org/10.1111/cogs.70064

Kirkham, S. (2025b). Scaling laws for nonlinear dynamical models of articulatory control. *JASA Express Letters*, *5*(2), 025201. https://doi.org/10.1121/10.0035560

Kirkham, S., & Strycharczuk, P. (2024). A dynamic neural field model of vowel diphthongisation. *Proceedings of the 13th International Seminar on Speech Production*.

Kramer, B. M., Stern, M. C., Wang, Y., Liu, Y., & Shaw, J. A. (2023). Synchrony and stability of articulatory landmarks in English and Mandarin CV sequences. *Proceedings of the 20th International Congress of Phonetic Sciences (ICPhS)*, 1022–1026.

Kröger, B. J., Schröder, G., & Opgen-Rhein, C. (1995). A gesture-based dynamic model describing articulatory movement data. *The Journal of the Acoustical Society of America*, *98*(4), 1878–1889. https://doi.org/10.1121/1.413374

Kuberski, S. R., & Gafos, A. I. (2023). How thresholding in segmentation affects the regression performance of the linear model. *JASA Express Letters*, *3*(9), 095202. https://doi.org/10.1121/10.0020815

Lee, D. N. (1998). Guiding Movement by Coupling Taus. *Ecological Psychology*, *10*(3–4), 221–250. https://doi.org/10.1080/10407413.1998.9652683

Liu, Y., Wang, Y., Stern, M. C., Kramer, B. M., & Shaw, J. A. (2023). Temporal scope of articulatory slowdown under informational focus: Data from English and Mandarin. *Proceedings of the 20th International Congress of Phonetic Sciences*, 1691–1695.

Löfqvist, A., & Gracco, V. L. (1997). Lip and Jaw Kinematics in Bilabial Stop Consonant Production. *Journal of Speech, Language, and Hearing Research*, *40*(4), 877–893. https://doi.org/10.1044/jslhr.4004.877

Moran, D. W., & Schwartz, A. B. (1999). Motor cortical representation of speed and direction during reaching. *Journal of Neurophysiology*, *82*(5), 2676–2692. https://doi.org/10.1152/jn.1999.82.5.2676

Mücke, D., Roessig, S., Thies, T., Hermes, A., & Mefferd, A. (2024). Challenges with the kinematic analysis of neurotypical and impaired speech: Measures and models. *Journal of Phonetics*, *102*, 101292. https://doi.org/10.1016/j.wocn.2023.101292

Munhall, K. G., Ostry, D. J., & Parush, A. (1985). Characteristics of velocity profiles of speech movements. *Journal of Experimental Psychology: Human Perception and Performance*, *11*(4), 457–474. https://doi.org/10.1037/0096-1523.11.4.457

Nam, H., & Saltzman, E. (2003). A Competitive, Coupled Oscillator Model of Syllable Structure. *Proceedings of the 15th International Congress of Phonetic Sciences*, 2253–2256.

Ostry, D. J., Cooke, J. D., & Munhall, K. G. (1987). Velocity curves of human arm and speech movements. *Experimental Brain Research*, *68*(1), 37–46. https://doi.org/10.1007/BF00255232

Ostry, D. J., Keller, E., & Parush, A. (1983). Similarities in the control of the speech articulators and the limbs: Kinematics of tongue dorsum movement in speech. *Journal of Experimental Psychology: Human Perception and Performance*, *9*(4), 622–636. https://doi.org/10.1037/0096-1523.9.4.622

Ostry, D. J., & Munhall, K. G. (1985). Control of rate and duration of speech movements. *The Journal of the Acoustical Society of America*, *77*(2), 640–648. https://doi.org/10.1121/1.391882

Parrell, B. (2011). Dynamical account of how /b, d, g/ differ from /p, t, k/ in Spanish: Evidence from labials. *Laboratory Phonology*, *2*(2), 423–449. https://doi.org/10.1515/labphon.2011.016

Parrell, B., Mefferd, A., Harper, S., Roessig, S., & Mücke, D. (2023). Using computational models to characterize the role of motor noise in speech: The case of amyotrophic lateral sclerosis. *Proceedings of the 20th International Congress of Phonetic Sciences*, 878–882.

Perkell, J. S., Matthies, M. L., Svirsky, M. A., & Jordan, M. I. (1993). Trading relations between tongue-body raising and lip rounding in production of the vowel /u/: A pilot "motor equivalence" study. *The Journal of the Acoustical Society of America*, *93*(5), 2948–2961. https://doi.org/10.1121/1.405814

Perrier, P., Abry, C., & Keller, E. (1988). Vers une modélisation des mouvements du dos de la langue. *Vers Une Modélisation Des Mouvements Du Dos de La Langue*, *2–1*, 45–63.

Pierrehumbert, J. B. (2001). Exemplar dynamics: Word frequency, lenition and contrast. In J. Bybee & P. J. Hopper (Eds.), *Frequency and the Emergence of Linguistic Structure* (pp. 137–158). John Benjamins.

Roessig, S., & Mücke, D. (2019). Modeling Dimensions of Prosodic Prominence. *Frontiers in Communication*, *4*(September), 1–19. https://doi.org/10.3389/fcomm.2019.00044

Roessig, S., Mücke, D., & Grice, M. (2019). The dynamics of intonation: Categorical and continuous variation in an attractor-based model. *PLoS ONE*, *14*(5). https://doi.org/10.1371/journal.pone.0216859

Roon, K. D., & Gafos, A. I. (2016). Perceiving while producing: Modeling the dynamics of phonological planning. *Journal of Memory and Language*, *89*, 222–243. https://doi.org/10.1016/j.jml.2016.01.005

Roon, K. D., Hoole, P., Zeroual, C., Du, S., & Gafos, A. I. (2021). Stiffness and articulatory overlap in Moroccan Arabic consonant clusters. *Laboratory Phonology: Journal of the Association for Laboratory Phonology*, *12*(1), 8. https://doi.org/10.5334/labphon.272

Saltzman, E. L., & Munhall, K. G. (1989). A Dynamical Approach to Gestural Patterning in Speech Production. *Ecological Psychology*, *1*(4), 333–382.

Schöner, G., Spencer, J., & Group, D. R. (2016). *Dynamic Thinking: A Primer on Dynamic Field Theory*. Oxford University Press.

Shaw, J. A. (2022). Micro-prosody. *Language and Linguistics Compass*, *16*(2), 1–21. https://doi.org/10.1111/lnc3.12449

Shaw, J. A. (2025). Unifying phonological and phonetic aspects of speech in dynamic neural fields: The case of laryngeal patterns in Japanese. *Phonological Studies*, *28*, 1–12.

Shaw, J. A., & Chen, W. (2019). Spatially Conditioned Speech Timing: Evidence and Implications. *Frontiers in Psychology*, *10*, 1–17. https://doi.org/10.3389/fpsyg.2019.02726

Shaw, J. A., & Tang, K. (2023). A dynamic neural field model of leaky prosody: Proof of concept. *Proceedings of the 2022 Annual Meeting on Phonology (AMP)*. https://doi.org/10.3765/amp.v10i0.5442

Sorensen, T., & Gafos, A. (2016). The Gesture as an Autonomous Nonlinear Dynamical System. *Ecological Psychology*, *28*(4), 188–215. https://doi.org/10.1080/10407413.2016.1230368

Stern, M. C. (2025). Dynamic Field Theory unifies discrete and continuous aspects of linguistic representations. *Proceedings of the Linguistic Society of America*, *10*(1). https://doi.org/10.3765/plsa.v10i1.5934

Stern, M. C., & Piñango, M. M. (2024). *Contextual modulation of language comprehension in a dynamic neural model of lexical meaning* (arXiv:2407.14701). arXiv. https://doi.org/10.48550/arXiv.2407.14701

Tiede, M. (2005). *MVIEW: Software for visualization and analysis of concurrently recorded movement data* [Computer software]. Haskins Laboratories.

Tilsen, S. (2016). Selection and coordination: The articulatory basis for the emergence of phonological structure. *Journal of Phonetics*, *55*, 53–77. https://doi.org/10.1016/j.wocn.2015.11.005

Tilsen, S. (2019). Motoric mechanisms for the emergence of non-local phonological patterns. *Frontiers in Psychology*, *10*(SEP), 1–25. https://doi.org/10.3389/fpsyg.2019.02143

Turk, A., & Shattuck-Hufnagel, S. (2020). *Speech Timing: Implications for Theories of Phonology, Phonetics, and Speech Motor Control*. Oxford University Press.

van Gelder, T. (1998). The dynamical hypothesis in cognitive science. *Behavioral and Brain Sciences*, *21*(5), 615–628. https://doi.org/10.1017/S0140525X98001733

Xu, Y. (1999). Effects of tone and focus on the formation and alignment of f0contours. *Journal of Phonetics*, *27*(1), 55–105. https://doi.org/10.1006/jpho.1999.0086

## 7. Appendix A: Derivation of second-order differential equation (Eq. 5)

$$\frac{d}{dt}((-x+T)/\dot{x}) = -r * (-x+T)/\dot{x}$$ (i) substitution

$$(-x+T) * \frac{d}{dt}(1/\dot{x}) + (1/\dot{x}) * \frac{d}{dt}(-x+T) = -r * (-x+T)/\dot{x}$$ (ii) product rule

$$(-x+T) * (-1/\dot{x}^2) * \ddot{x} + (1/\dot{x}) * \frac{d}{dt}(-x+T) = -r * (-x+T)/\dot{x}$$ (iii) chain rule, power rule

$$(-x+T) * (-1/\dot{x}^2) * \ddot{x} + (1/\dot{x}) * \left(-\dot{x} + \frac{dT}{dt}\right) = -r * (-x+T)/\dot{x}$$ (iv) sum rule

$$(-x+T) * (-1/\dot{x}^2) * \ddot{x} - 1 = -r * (-x+T)/\dot{x}$$ (v) $\frac{dT}{dt} = 0$

$$-\ddot{x}(-x+T)/\dot{x}^2 = -r * (-x+T)/\dot{x} + 1$$ (vi) simplification

$$-\ddot{x}(-x+T) = -r\dot{x} * (-x+T) + \dot{x}^2$$ (vii) simplification

$$\ddot{x} = r\dot{x} - \dot{x}^2/(-x+T)$$ (viii) simplification

## 8. Appendix B: Stimulus sentences

| Item # | Prominence | Stimulus | |
|---|---|---|---|
| 1 | Not prominent | Prompt | 雷峰塔必须重修还是拆除?<br>Lei2 feng1 ta3 bi4xu1 cho2ngxiu1 hai2shi4 chai1chu2?<br>‘Must Leifeng Pagoda be rebuilt or demolished?’ |
| | | Target | 雷锋塔必须重修，不是必须拆除。<br>Lei2feng1 Ta3 bi4xu1 **chong2xiu1**, bu2shi4 bi4xu1 chai1chu2.<br>‘Leifeng Pagoda must be rebuilt, not demolished.’ |
| | Prominent | Prompt | 雷峰塔必须重修还是可以重修?<br>Lei2 fe1ng ta3 bi4xu1 chong2xiu1 hai2shi4 ke3yi3 chong2xiu1?<br>‘Is it the case that Leifeng Pagoda must be rebuilt or (just) that it can be rebuilt?’ |
| | | Target | 雷锋塔必须重修，不是可以重修。<br>Lei2feng1 Ta3 **bi4xu1** chong2xiu1, bu2shi4 ke3yi3 chong2xiu1.<br>‘Leifeng Pagoda must be rebuilt, not (just) can be rebuilt.’ |
| 2 | Not prominent | Prompt | 我要打蜜蜂的翅膀还是触角?<br>Wo3 yao4 da3 mi4feng1 de0 chi4bang3 hai2shi4 chu4jiao3?<br>‘Should I hit the bee’s wing or antenna?’ |
| | | Target | 你要做的是打蜜蜂的翅膀，不是打蜜蜂的触角。<br>Ni3 yao4 zuo4 de0 shi4 da3 mi4feng1 de0 **chi4bang3**, bu2shi4 da3 mi4feng1 de0 chu4jiao3.<br>‘What you should do is to hit the bee’s wing, not to hit the bee’s antenna.’ |
| | Prominent | Prompt | 我要打蜜蜂的翅膀还是蝴蝶的翅膀?<br>Wo3 yao4 da3 mi4feng1 deo chi4bang3 hai2shi4 hu2die2 de0 chi4bang3?<br>‘Should I hit the bee’s wing or the butterfly’s wing?’ |
| | | Target | 你要做的是打蜜蜂的翅膀，不是打蝴蝶的翅膀。<br>Ni3 yao4 zuo4 de0 shi4 da3 **mi4feng1** de0 chi4bang3, bu2shi4 da3 hu2die2 de0 ch14bang3.<br>‘What you should do is to hit the bee’s wing, not to hit the butterfly’s wing.’ |
| 3 | Not prominent | Prompt | 这张卡必须留着还是扔掉?<br>Zhe4 zhang1 ka3 bi4xu1 liu2zhe0 hai2shi4 reng1 diao4?<br>‘Must this card be kept or thrown away?’ |
| | | Target | 这张卡必须留着，不是必须扔掉。<br>Zhe4 zhang1 ka3 bi4xu1 **liu2**-zhe0, bu2shi4 bi4xu1 reng1diao4.<br>‘This card must be kept, not thrown away.’ |
| | Prominent | Prompt | 这张卡必须留着还是可以留着?<br>Zhe4 zhang1 ka3 bi4xu1 liu2zhe0 hai2shi4 ke3yi3 liu2zhe0?<br>‘Is it the case that this card must be kept or (just) that it can be kept?’ |
| | | Target | 这张卡必须留着，不是可以留着。<br>Zhe4 zhang1 ka3 **bi4xu1** liu2-zhe0, bu2shi4 ke3yi3 liu2-zhe0.<br>‘This card must be kept, not (just) can be kept.’ |
| 4 | Not prominent | Prompt | 我要把银行卡密码告诉她还是告诉你?<br>Wo3 yao4 ba3 yin2hang2 ka3 mi4ma3 gao4su4 ta1 hai2shi4 gao4su4 ni3? |
| | | | 'Should I tell her the credit card password, or should I tell you?' |
| | | Target | 你要做的是把银行卡密码告诉她，不是把银行卡密码告诉我。<br>Ni3 yao4 zuo4 de0 shi4 ba3 yin2hang2 ka3 mi4 ma3 gao4su4 **ta1**, bu2shi4 ba3 yin2hang2 ka3 mi4 ma gao4su4 wo3.<br>'What you should do is to tell her the credit card password, not to tell me the credit card password.' |
| | Prominent | Prompt | 我要把银行密码还是银行卡号告诉她?<br>Wo3 yao4 ba3 yin2hang2 mi4ma3 hai2shi4 yin2hang2 ka3hao4 gao4su4 ta1?<br>'Should I tell her the credit card password or the credit card number?' |
| | | Target | 你要做的是把银行卡密码告诉她，不是把银行卡号告诉她。<br>Ni3 yao4 zuo4 de0 shi4 ba3 yin2hang2 ka3 **mi4** ma3 gao4su4 ta1, bu2shi4 ba3 yin2hang3 ka3 hao4 gao4su4 ta1.<br>'What you should do is to tell her the credit card password, not the credit card number.' |
| 5 | Not prominent | Prompt | 这是我爸比的咖啡还是茶?<br>Zhe4 shi4 wo3 ba4 bi3 de0 ka1fei1 hai2shi4 cha2?<br>'Is this my daddy's coffee or tea?' |
| | | Target | 这是你爸比的咖啡，不是你爸比的茶。<br>Zhe4 shi4 ni3 ba4bi0 de0 **ka1fei1**, bu2shi4 ni3 ba4bi0 de0 cha2.<br>'This is your daddy's coffee, not your daddy's tea.' |
| | Prominent | Prompt | 这是我爸比的咖啡还是我妈咪的咖啡?<br>Zhe4 shi4 wo3 ba4 bi3 de0 ka1fei1 hai2shi4 wo3 ma1 mi1 de0 ka1fei1?<br>'Is this my daddy's coffee or my mommy's coffee?' |
| | | Target | 这是你爸比的咖啡，不是你妈咪的咖啡。<br>Zhe4 shi4 ni3 **ba4bi0** de0 ka1fei1, bu2shi4 ni3 ma1mi0 de0 ka1fei1.<br>'This is your daddy's coffee, not your mommy's coffee.' |
| 6 | Not prominent | Prompt | 我应该骂他还是骂你?<br>Wo3 ying1gai1 ma4 ta1 hai2shi4 ma4 ni3?<br>'Should I scold him or scold you?' |
| | | Target | 你骂他就行了，别骂我。 |
| | | | Ni3 ma4 **ta1** jiu4 xing2 le0, bie2 ma4 wo3.<br>‘Just scold him; don’t scold me.’ |
| | Prominent | Prompt | 我应该骂他还是打他?<br>Wo3 ying1gai1 ma4 ta1 hai2shi4 da3 ta1?<br>‘Should I scold him or hit him?’ |
| | | Target | 你骂他就行了，别打他。<br>Ni3 **ma4** ta1 jiu4 xing2 le0, bie2 da3 ta1.<br>‘Just scold him; don’t hit  him.’ |
| 7 | Not prominent | Prompt | 我们三个里爸比最高还是最矮?<br>Wo3men0 san1 ge4 li3 ba4 bi3 zui4gao1 hai2shi4 zui4 a3?<br>‘Is daddy the tallest or the shortest among us three?’ |
| | | Target | 我们三个里爸比最高，不是爸比最矮。<br>Wo3men0 san1ge4 li3 ba4bi0 zui4 **gao1**, bu2shi4 ba4bi0 zui4 ai3.<br>‘Daddy is the tallest among us, not the shortest.’ |
| | Prominent | Prompt | 我们三个里爸比最高还是妈咪最高?<br>Wo3men0 san1 ge4 li3 ba4 bi3 zui4gao1 hai2shi4 ma1 mi1 zui4gao1?<br>‘Is daddy the tallest  among us three, or is mommy the tallest?’ |
| | | Target | 我们三个里爸比最高，不是妈咪最高。<br>Wo3men0 san1ge4 li3 **ba4bi0** zui4 gao1, bu2shi4 ma1mi0 zui4 gao1.<br>‘Daddy is the tallest among us, not mommy.’ |
| 8 | Not prominent | Prompt | 在学校里可以骂同学吗？骂老师呢?<br>Zai4 xue2xiao4 li3 ke3yi3 ma4 tong2xue2 ma? Ma4 lao3shi1 ne?<br>‘Is it okay to scold your classmate at school? And how about scolding your teacher?’ |
| | | Target | 在学校里骂同学可以，但是骂老师不行。<br>Zai4 xue2xiao4 li3 ma4 **tong2xue2** ke3yi3, dan4shi4 ma4 lao3shi1 bu4 xing2.<br>‘It is okay to scold your classmate at school, but it is not okay to scold your teacher.’ |
| | Prominent | Prompt | 在学校里可以骂同学吗？打同学呢? |
| | | | Zai4 xue2xiao4 li3 ke3yi3 ma4 tong2xue2 ma? Da3 tong2xue2 ne0?<br>‘Is it okay to scold your classmate at school? And how about hitting your classmate?’ |
| | | Target | 在学校里骂同学可以，但打同学不行。<br>Zai4 xue2xiao4 li3 **ma4** tong2xue2 ke3yi3, dan4shi4 da3 tong2xue2 bu4 xing2.<br>‘It is okay to scold your classmate at school, but it is not okay to hit your classmate.’ |

Table B.1: Mandarin stimuli. Target CV sequences are underlined; prominent words are bolded.

| Item # | Prominence | Stimulus | |
|---|---|---|---|
| 9 | Not prominent | Prompt | Did he see meaning in the picture? |
| | | Target | He saw meaning in the **book**, not in the picture. |
| | Prominent | Prompt | Did he see chaos in the picture? |
| | | Target | He saw **meaning** in the picture, not chaos. |
| 10 | Not prominent | Prompt | Did she see you by the beach? |
| | | Target | She saw me by the **river**, not by the beach. |
| | Prominent | Prompt | Did she see Thomas by the river? |
| | | Target | She saw **me** by the river, not Thomas. |
| 11 | Not prominent | Prompt | Did you see beaded necklaces? |
| | | Target | I saw beaded **bracelets**, not beaded necklaces. |
| | Prominent | Prompt | Did you see woven necklaces? |
| | | Target | I saw **beaded** necklaces, not woven ones. |
| 12 | Not prominent | Prompt | Do you hear the robot's jaw beep loudly? |
| | | Target | I hear the robot's jaw beep **quietly**, not loudly. |
| | Prominent | Prompt | Do you hear the robot's jaw screech? |
| | | Target | I hear the robot's jaw **beep**, not screech. |
| 13 | Not prominent | Prompt | Are you talking about the knee body circumference? |
| | | Target | I'm talking about the knee body **radius**, not the knee body circumference. |
| | Prominent | Prompt | Are you talking about the knee ligament? |
| | | Target | I'm talking about the knee **body**, not the knee ligament. |
| 14 | Not prominent | Prompt | Is she a knee model client? |
| | | Target | She's a knee model **representative**, not a knee model client. |
| | Prominent | Prompt | Is she a knee surgeon? |
| | | Target | She's a knee **model**, not a knee surgeon. |
| 15 | Not prominent | Prompt | Does he frequently bother the customers? |
| | | Target | He frequently bothers the **workers**, not the customers. |
| | Prominent | Prompt | Does he frequently help the customers? |
| | | Target | He frequently **bothers** the customers; he doesn't frequently help them. |
| 16 | Not prominent | Prompt | Does she frequently model shoes? |
| | | Target | She frequently models **shirts**, not shoes. |
| | Prominent | Prompt | Does she frequently sell shoes? |
| | | Target | She frequently **models** shoes; she doesn't frequently sell them. |

Table B.2: English stimuli. Target CV sequences are underlined; prominent words are bolded.